\documentclass[journal]{IEEEtran}

\usepackage[pdftex]{graphicx}
\usepackage{subfigure}
\usepackage{amssymb}
\usepackage{latexsym}
\usepackage{cite}
\usepackage{hyperref}
\usepackage{bm}

\newcommand {\comment}[1] {}
\newcommand {\comments}[1] {}

\newlength \figwidth
\if@twocolumn
\else
\fi
\def\figratio1{.1}

\begin{document}

\title{Rotation Invariant Descriptors for Galaxy Morphological Classification}

\author{Hubert Cecotti
\thanks{H. Cecotti is with the Department of Computer Science, College of Science and Mathematics, Fresno State University, Fresno, Ca, USA.}}

\maketitle

\begin{abstract}
The detection of objects that are multi-oriented is a difficult pattern recognition problem. In this paper, we propose to evaluate the performance of different families of descriptors for the classification of galaxy morphologies. We investigate the performance of the Hu moments, Flusser moments, Zernike moments, Fourier-Mellin moments, and ring projection techniques based on 1D moment and the Fourier transform. We consider two main datasets for the performance evaluation. The first dataset is an artificial dataset based on representative templates from 11 types of galaxies, which are evaluated with different transformations (noise, smoothing), alone or combined. The evaluation is based on image retrieval performance to estimate the robustness of the rotation invariant descriptors with this type of images. The second dataset is composed of real images extracted from the Galaxy Zoo 2 project. The binary classification of elliptical and spiral galaxies is achieved with pre-processing steps including morphological filtering and a Laplacian pyramid. For the binary classification, we compare the different set of features with Support Vector Machines, Extreme Learning Machine, and different types of linear discriminant analysis techniques. The results support the conclusion that the proposed framework for the binary classification of elliptical and spiral galaxies provides an area under the receiver operating characteristic curve reaching 99.54\%, proving the robustness of the approach for helping astronomers to study galaxies. 
\end{abstract}


\IEEEpeerreviewmaketitle

\section{Introduction}

Pattern recognition and machine learning techniques are now reaching the stars through the use of advanced techniques to classify galaxy morphologies~\cite{ball2010}. Given the different shapes and orientations of the galaxies, robust descriptors are required to classify them. In particular, these descriptors should be translation, scale, and rotation invariant when applied in large collection of images. The latter characteristic has been a key problem since the early days of pattern recognition. In this study, we propose to analyze the performance of rotation invariant techniques to classify different morphologies of galaxies. The morphological classification of galaxies has been typically done visually as this difficult task requires some prior experience and knowledge with the type of images to analyze. In the last years, some projects such as the series of Galaxy Zoo projects~\cite{2011MNRAS.410..166L,willett2013}, have significantly enhanced the classification of galaxy morphologies. These projects have shown that large datasets of galaxy images can be analyzed by non-scientist volunteers, and combining the analysis across multiple participants can provide some reliable measurements despite the inner subjective evaluations. However, the substantial increase of images obtained from telescopes cannot be matched by the distributed manual efforts. It is therefore necessary to provide techniques based on machine learning and image processing to classify images of galaxies~\cite{shamir2009}. Furthermore, classifiers can be directly used as a means to rank human performance for labeling as the performance is directly linked to the level of noise in the data~\cite{shamir2016}. For instance, an average number of 44 users analyzed each galaxy to determine its shape (e.g., smooth and rounded or not). A key issue related to manual labeling is the level of agreement among the different participants. With only about 8\% of the galaxies that were classified when the agreement among the voters was superior or equal to 95\%, it is impossible to estimate with confidence all the remaining galaxies~\cite{willett2013}. These results illustrate both the difficulty of the task and the need of more reliable methods.

\begin{figure}[!t]
\centering
\includegraphics[width=\figwidth]{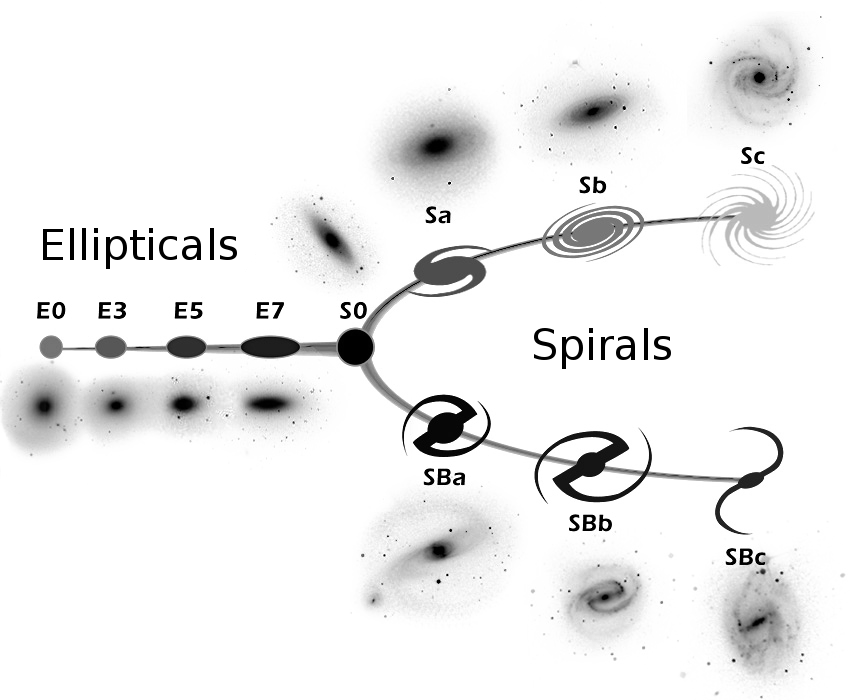}
\caption{Hubble's classification sequence.}
\label{fig:hubble}
\end{figure}

Galaxies can be divided into three main classes corresponding to their shape: spirals (disk dominated shape) (S), elliptical (spheroidal-looking) (E), and irregulars (I). The description of the shapes started in the 19th century with the notion of spirals from William Parsons, 3rd Earl of Rosse. This classification system is called the Hubble sequence~\cite{hubble1926}. The morphology of galaxies encodes information related to the orbital parameters, and assembly history of galaxies can be decoded through the analysis of their morphology. Their content includes gas, dust, stars, and the central black hole. In addition, the morphology is closely related to the local environment of the galaxy because mutual interactions like tides, shocks in cluster environments, and direct mergers can all modify the shape of the galaxy's gravitational potential~\cite{buta2001,buta2011}. The tuning fork diagram is an arrangement of galaxies according to their rotation. In this diagram, it starts with elliptical galaxies (E) then forks into two types of spirals: with (SB) and without (S) a central bar-shaped structure (see Fig.~\ref{fig:hubble}). Elliptical galaxies can be subsequently decomposed in relation to their degree of ellipticity in the sky. $Ex$ corresponds to an elleptical galaxy with $x=10(1−b/a)$ for an ellipse with semi-major and semi-minor axes of lengths $a$ and $b$, respectively. In the subsequence sections, we will consider E0, E3, and E7. Spiral galaxies are typically described as a flat rotating disk, which contains stars, gas, dust, and a central concentration of stars. They can be decomposed in relation to the tightness of their spiral arms. A lower-case letter is added to the name of the class to determine the spiral structure appearance, e.g., Sa/SBa for tightly wound, smooth arms; large, bright central bulge; Sb/SBb for less tightly wound spiral arms than Sa/SBa; Sc/SBc for loosely wound spiral arms. In 1936, Hubble revised his classification system to include a fourth major galaxy class: S0 (lenticular) galaxies that were armless disk galaxies. They represent the transition from ellipticals to fully developed spirals. This type remains a research question due to its relationship to spiral and elliptical types. The S0 class is between E and S in the sequence. The Hubble sequence has been extended through the de Vaucouleurs system, which is a finer description of the galaxy morphologies, introducing features such as the presence of a nuclear bar~\cite{vaucouleurs1959}.

The goal of this paper is multifold: 1) to provide a comprehensive description of current state of the art rotation invariant descriptors, 2) to analyze the performance of these different sets of descriptors on multiple datasets (artificial and real) with different level of noise, corresponding the galaxy morphologies, 3) to propose a framework for the classification of galaxies using rotation invariant descriptors. 
The remainder of the paper is organized as follows. First, related works and the various techniques to extract rotation invariant features are presented in Sections~\ref{sec:related} and~\ref{sec:methods}. The artificial datasets, the real images, and the proposed preprocessing steps for denoising the images are detailed in Section~\ref{sec:datasets}. The performance of the different approaches are then presented in Section~\ref{sec:results}. Finally, the impact of the results are discussed in Section~\ref{sec:discussion}.

\section{Related works}
\label{sec:related}

The classification of galaxy morphologies is a difficult and subjective task that requires an expert or a committee of experts to label images such as what was used in the Galaxy Zoo projects. It is difficult as the details that can be observed separating two morphologies can be subtle, requiring the eye of an experienced observer to identify faint objects. 
Hence, the creation of the ground truth is subjective, leading to different observers assigning galaxies to different classes. In addition, it is a challenging task at multiple levels for the creation of a training dataset. First, the images must be segmented in relation to the apparent radius of the object. For instance, the radius of the object can be estimated through the Petrosian radius, which is a distance-independent measurements of galaxy profiles~\cite{petrosian1976}. A galaxy can be described through its S\'ersic~\cite{sersic1963} or Jaffe~\cite{jaffe1983} profile to describe the light distribution and have been used to determine galaxy morphologies~\cite{vanderWel2008}. Typical global binarization techniques that can be used for symbol detection in technical documents cannot be applied directly given the continuity between the shape and its background with images of galaxies. Multiple approaches have been proposed.
Neural networks using backpropagation have first been used~\cite{naim1995}, then decision trees~\cite{owens1996}. In~\cite{bazell2001}, they compared a Naive Bayes classifier, an artificial neural network, and a decision tree using a sample of 800 galaxies, showing the interest of ensembles of classifiers. In~\cite{delacalleja2004}, they used a neural network, and a locally weighted regression method, with an ensembles of classifiers, and obtained 91\% accuracy when considering E, S and I galaxy classes. The geometric shape features and direct pixel images of galaxies have been compared and classified with neural networks~\cite{goderya2002}, highlighting the interest of shape descriptors. Ganalyzer was proposed as a tool for automatic galaxy image analysis~\cite{shamir2011}. An image analysis unsupervised learning algorithm using a weighted Euclidean distance was proposed for the detection of peculiar galaxies~\cite{shamir2012}. State of the art performance has been obtained in~\cite{dieleman2015} thanks to data augmentation, regularization, parameter sharing, and model averaging, highlighting the performance of convolutional neural networks (CNNs), which is another type of feedforward artificial neural network.

\section{Methods}
\label{sec:methods}

The problem related to the classification of objects that can be oriented in different angles can be treated in different ways. First, the problem can be simply ignored and the classifier will have to deal with features corresponding to different orientations. In such a case, considering a discriminant approach, a deep architecture should be considered to achieve a type of ``or'' between the different possible orientations. This approach can be judicious if there are only a few types of angles for the rotations and if these angles are equally distributed between the training and the test stage. Then, the images can be clustered in relation to the different angles. With density based approach and a large number of labeled examples covering all the different orientations, the different orientations may be ignored. The second approach is based on the estimation of the orientation of the image to reorient the images in relation to their main direction. If there is no main direction or if its estimation is difficult, it can add errors to the following processing steps. In the third way, descriptors invariant to the rotation can be extracted, i.e., the descriptors of an object will be identical, independently of the object's orientation.  

A key approach for the extraction of rotation invariant features is to transform the input image, which is originally in Cartesian coordinates into polar coordinates, changing the rotation invariant problem into a circular-shift invariant problem, where features can be obtained from the whole image (2D) or from the different rings (1D) that compose the image. Before this step, the gravity center of the chosen image must be estimated and the image should be centered on its gravity center. We consider the discrete representation of an image $f_c$ of size $N_x \times N_y$ described in Cartesian coordinates by $f_c(x,y)$ with $0 < x \leq N_x$, and $0 < y \leq N_y$. The image in polar coordinates is described by $f_p(\rho,\theta)$ with $0 < \rho \leq R_{max}$, $R_{max} \in \mathbb{R}^{\*}$, and $0 \leq \theta < 2\pi$. The discrete description of the image in polar coordinate $f_p(r,t)$ is a matrix of size $N_{\rho} \times N_{\theta}$, we have $0 <  r \leq N_{\rho}$ and $0 < t \leq N_{\theta}$ with the angular and radial sampling steps defined by: $\Delta \theta = 2\pi/N_{\theta}$, $\Delta\rho=R_{max}/N_{\rho}$, and $\rho = r \Delta \rho$ and $\theta = t \Delta\theta$. It has been shown that if normalized invariant moments in circular windows are used, then template matching in rotated images becomes similar to template matching in translated images~\cite{goshtasby1985}. 

\subsection{Flusser and Hu moments}

Moment invariants have been introduced for pattern recognition problems by Hu~\cite{hu1962}. They have been successfully used in a large number of 2D shape detection problems. These moments have been further analyzed and developed as complex moments~\cite{flusser2000}. A complex moment $c_{pq}$ of the order $(p+q)$ of an integrable image function $f_c(x, y)$ is defined by:
\begin{eqnarray}
c_{pq} & = & \int\limits_{-\infty}^{+\infty} \int\limits_{-\infty}^{+\infty} (x+iy)^p(x-iy)^q f_c(x,y) dxdy
\end{eqnarray}
with $i$ denoting the imaginary unit ($\sqrt{-1}$).
The complex moments can be represented through geometric moments $m_{pq}$:
\begin{eqnarray}
c_{pq} & = \sum\limits_{k=0}^{p} \sum\limits_{j=0}^{q} & \left( \begin{tabular}{c}p\\k\end{tabular} \right)  \left( \begin{tabular}{c}q\\j\end{tabular} \right) \cdot \\
       &   & (-1)^{q-j} i^{p+q-k-j} m_{k+j,p+q-k-j} \nonumber
\end{eqnarray}
where the two-dimensional geometric moment of order $p+q$ is defined by:
\begin{eqnarray}
m_{pq} & = & \int\limits_{-\infty}^{+\infty} \int\limits_{-\infty}^{+\infty} x^p y^q f_c(x,y) dxdy
\end{eqnarray}

After the rotation of the image by an angle $\beta$, we obtain:
\begin{eqnarray}
c_{pq}^{rot} & = & c_{pq} \cdot e^{-i(p-q)\beta} 
\end{eqnarray}

The Flusser moments are defined as follows (second and third orders: $\psi_1$ to $\psi_6$, fourth order $\psi_7$ to $\psi_{11}$) 
\begin{eqnarray}  
\begin{tabular}{ll}
$\psi_1 = c_{11}$ & $\psi_2 = c_{21}c_{12}$ \\
$\psi_3 = Re(c_{20}c_{12}^2)$ & $\psi_4 = Im(c_{20}c_{12}^2)$ \\
$\psi_5 = Re(c_{30}c_{12}^3)$ & $\psi_6 = Im(c_{30}c_{12}^3)$ \\
$\psi_7 = c_{22}$ & $\psi_8 = Re(c_{31}c_{12}^2)$ \\
$\psi_9 = Im(c_{31}c_{12}^2)$ & $\psi_{10} =  Re(c_{40}c_{12}^4)$ \\
$\psi_{11} =  Im(c_{40}c_{12}^4)$ & 
\end{tabular}
\end{eqnarray}
It is worth mentioning that low-order moments are less sensitive to noise than the higher-order ones. 
The Flusser invariants are denoted by:
\begin{eqnarray}
F_{Flusser} & = & \{ \psi_{1},\ldots,\psi_{11} \}
\end{eqnarray}

The Hu invariants~\cite{hu1962} can be defined in relation to the complex moments described before~\cite{flusser2000}.
\begin{eqnarray}
\begin{tabular}{ll}
$\phi_1 = c_{11}$ & $\phi_2 = c_{20}c_{02}$ \\
$\phi_3 = c_{30}c_{03}$ & $\phi_4 = c_{21}c_{12}$ \\
$\phi_5 = Re(c_{30}c_{12}^3)$ & $\phi_6 = Re(c_{20}c_{12}^2)$ \\ 
$\phi_7 = Im(c_{30}c_{12}^3)$ &  
\end{tabular}
\end{eqnarray}

The Hu invariants are denoted by:
\begin{eqnarray}
F_{Hu} & = & \{ \phi_1,\ldots,\phi_7 \}
\end{eqnarray}

\subsection{Zernike rotation invariant}

Zernike provided a set of complex polynomials that form a complete orthogonal set over the interior of the circle of a radius equal to 1, ($x^2+y^2=1$). The set of these polynomials is denoted by $V = \{ V_{n,m}(r,\theta) \}$~\cite{zernike1934,khotanzad1990}.
These polynomials are defined by:
\begin{eqnarray}
V_{n,m}(r,\theta) & = & R_{n,m}(r) e^{im\theta}
\end{eqnarray}
where $n \in \mathbb{N}_0$, $m \in \mathbb{Z}$, with $|m|\leq n$ and $(n-|m|) mod 2 =0$, $r$ represents the length of the vector $\vec{OP}$ from the the origin O to P $(r \cdot cos(\theta),r \cdot sin(\theta))$, with $\theta \in [0,2\pi]$.

The radial polynomial $R_{n,m}$ is defined by:
\begin{eqnarray}
R_{n,m}(r) & = & \sum\limits_{k=0}^{(n-|m|)/2} (-1)^k \cdot r^{n-2k} \cdot \\
          &    &\frac{(n-k)!}{k!(\frac{n+|m|}{2}-k)!(\frac{n-|m|}{2}-k)!}  \nonumber 
\end{eqnarray}
with $R_{n,m}(r)=R_{n,-m}(r)$.

Given the relationships between $n$ and $m$, we define two vectors $N_z$ and $M_z$ containing a subset of the possible values for $n$ and $m$. Hence, $R_{n,m}$ can be estimated for each couple: $R_{n,m}=R(j)$, and $V_{n,m}(r,\theta)=V_{j}(r,\theta)$ with $n=N_z(j)$ and $m=M_z(j)$ can be estimated in relation to $j$ only. 

For Zernike moments of order $n$, we have $(n+1)(n+2)/2$ different polynomials. For instance, for $n=5$, we can consider the following vectors:
\begin{eqnarray}
N_z & = & [0,  1,  1,  2,  2,  2,  3,  3,  3,  3,  4,\\
    & = &  4,  4,  4,  4,  5,  5,   5,  5,  5,  5] \nonumber  \\
M_z & = & [0, -1,  1, -2,  0,  2, -3, -1,  1,  3, -4,\nonumber  \\
    &  &  -2,  0,  2,  4,  -5, -3, -1, 1,  3,  5] \nonumber 
\end{eqnarray}

Zernike moments of order $n$ with repetition $m$, $A_{n,m}$ or $A_j$, based on the projection of the image onto the orthogonal basis functions, are defined by:
\begin{eqnarray}
A_{n,m} & = & \sum\limits_{r=1}^{N_{\rho}} \sum\limits_{t=1}^{N_{\theta}} f_p(r,t)  V_{n,m}^*(r \cdot \Delta \rho,t \cdot \Delta \theta) \\
A_j & = & \sum\limits_{r=1}^{N_{\rho}} \sum\limits_{t=1}^{N_{\theta}} f_p(r,t)  V_{j}^*(r \cdot \Delta \rho,t \cdot \Delta \theta) \nonumber
\end{eqnarray}
with $A_{n,m}^*=A_{n,-m}$.

Following a rotation of angle $\alpha$, 
\begin{eqnarray}
A_{n,m}^{rot} & = & A_{n,m} \cdot e^{-im\alpha}
\end{eqnarray}
Therefore, $|A_{n,m}|=|A_{n,-m}|$ because $A_{n,m}^*=A_{n,-m}$. So there is only the need to compute the $A_{n,m}$ for $m \geq 0$. $N_z$ and $M_z$ for $n=5$ becomes:
\begin{eqnarray}
N_z & = & [0, 1, 2, 2, 3, 3, 4, 4, 4, 5, 5, 5] \\ 
M_z & = & [0, 1, 0, 2, 1, 3, 0, 2, 4, 1, 3, 5] \nonumber 
\end{eqnarray}

The set of rotation invariant features becomes:
\begin{eqnarray}
 F_{zernike} & = & \{|A_j|\} ~\mbox{with}~ M_z(j) \geq 0
\end{eqnarray}

\subsection{Ring projection}

The image is first centered on its gravity center and transformed into polar coordinate in relation to a maximum radius~\cite{tang1991}. Each ring can be defined by its first raw moment (the mean $\mu$) and its higher central moments (variance $\sigma$, skewness $\gamma$, and kurtosis $\kappa$), defining a set of rotation invariant features:
\begin{eqnarray}
\mu(r) & = & \frac{1}{N_{\theta}} \sum\limits_{t=1}^{N_{\theta}} f_c(r,t) \\
\sigma(r) & = & \frac{1}{N_{\theta}} \sum\limits_{t=1}^{N_{\theta}} (f_c(r,t)-\mu(r))^2 \nonumber  \\
\gamma(r) & = & \frac{\mu(r)^3}{\sigma(r)^3} \nonumber  \\
\kappa(r) & = & \frac{\mu(r)^4}{\sigma(r)^4} \nonumber 
\end{eqnarray}

The set of rotation invariant features becomes:
\begin{eqnarray}
F_{ring} & = & \left[ \begin{tabular}{ccc} $\mu(1)$ & $\ldots$ & $\mu(N_{\rho})$ \\
																					 $\sigma(1)$ & $\ldots$ & $\sigma(N_{\rho})$ \\
																					 $\gamma(1)$ & $\ldots$ & $\gamma(N_{\rho})$ \\
																					$\kappa(1)$ & $\ldots$ & $\kappa(N_{\rho})$ 
											\end{tabular} \right] 
\end{eqnarray}

\subsection{FFT based rotation invariant}

This approach is also based on the previous ring projection technique. In each ring, we consider the magnitude of the Fourier coefficient, which are shift invariant.
\begin{eqnarray}
Y(r,k) & = & \sum\limits_{t=1}^{N_{\theta}} f_c(r,t) \cdot e^{-i2\pi k n \ N_{\theta} }
\end{eqnarray}
In order to use the Fast Fourier Transform, we consider $N_{\theta}$ as a power of 2.
The set of rotation invariant features can be:
\begin{eqnarray}
F_{fft} & = & |Y(r,1..N_{\theta}/2+1)| 
\end{eqnarray}
with $0 <  r \leq N_{\rho}$. In order to be less sensitive to the high frequencies, we consider a log selection of the different features. We select the values for $|Y(r,1)|$, $|Y(r,2)|$, $|Y(r,3)|$, and the mean (bandpower) between $|Y(r,2+2^{(p-1)})|$ and $|Y(r,1+2^{p})|$ with $1<p \leq log_2(N_{\theta}/2)$. It leads to $r(log_2(N_{\theta}/2)+2)$ features.

\subsection{Fourier Mellin transform invariant}

The Mellin transform and the Fourier-Mellin transform (FMT) are widely used transforms in pattern recognition with the motivation to extract features invariant to rotation and scale~\cite{grace1991,sheng1991}. This transform has been successfully used with learning vector quantization and the K-nearest neighbors for character and symbol classification in technical documents~\cite{adam2000,adam2000a}. The analytical FMT of an image in polar coordinate $f_p$ is defined by:
\begin{eqnarray}
M_{f\sigma}(k,v) & = & \frac{1}{2\pi} \int\limits_0^{\infty} \int\limits_0^{2\pi} f_p(\rho,\theta) r^{\sigma-iv} e^{-ik \theta} d\theta \frac{dr}{r}
\end{eqnarray}
with $\forall (k,v) \in \mathbb{Z} \times \mathbb{R}$ and $\sigma>0$.

If we consider the image $g_p$ corresponding to the rotation of an angle $\alpha$ and scale change of factor $\beta$ of the object described in $f_p$, we have $g_p(\rho,\theta)=f_p(\alpha \rho,\beta+\theta)$. As the two images contain the same shape, the background being set to 0, the analytical FMT of $g_p$ in relation to $f_p$ can be expressed as: 
\begin{eqnarray}
M_{g\sigma}(k,v) & = & \frac{1}{2\pi} \int\limits_0^{\infty} \int\limits_0^{2\pi} f_p(\alpha\rho,\beta+\theta) \cdot \\ 
                 &   &    r^{\sigma-iv} e^{-ik \theta} d\theta \frac{dr}{r} \nonumber \\
                 & = & \alpha^{-sigma+iv} e^{ik\beta} M_{f\sigma}(k,v) 
\end{eqnarray}
The FMT descriptors of two similar objects presented in two images with different orientation will only differ by a phase factor. Therefore, a set of descriptors invariant to the rotation can be obtained by selecting the magnitude of the FMT descriptors~\cite{casasent1976}.

The discrete FMT approximation $M_{f\sigma}(k,v)$ is estimated for $(k,v) \in [-K,K] \times [-V,V]$. The parameter $\sigma$ is set to 0.5~\cite{goh1985}. The sampling step over $v$ is set to 1 and the approximation is given by:
\begin{eqnarray}
M^p_{f\sigma}(k,v) & = & \Delta\rho\Delta\theta \sum\limits_{j=1}^{N_\rho} F_k(\rho) (\rho)^{\sigma-iv-1}  \\
F_k(\rho) & = & \sum\limits_{j=1} f_p(\rho,\theta) e^{-ikj/N_{\theta}}
\end{eqnarray}

In addition, for real-valued functions such as images, the analytical FMT is symmetrical: 
\begin{eqnarray}
M_{f\sigma}(-k,-v) & = & \overline{M_{f\sigma}(k,v)}
\end{eqnarray}
Therefore, the FMT can be estimated for close to half of the elements in $(k,v)$, such as $[-K,K] \times [0,V]$.
We go from:
\begin{eqnarray}
\left[ \begin{tabular}{ccc}    $M_{f\sigma}(-k,-v)$ & $M_{f\sigma}(-k,0)$ & $M_{f\sigma}(-k,v)$ \\
															 $M_{f\sigma}(0,-v)$ & $M_{f\sigma}(0,0)$ & $M_{f\sigma}(0,v)$ \\
															 $M_{f\sigma}(k,-v)$ & $M_{f\sigma}(k,0)$ & $M_{f\sigma}(k,v)$
                       \end{tabular} \right]
\end{eqnarray}
with $(k,v)  \in [1,K] \times [1,V]$ to the estimation of only:
\begin{eqnarray}
\left[ \begin{tabular}{ccc}    -                 & -                & -                \\
															 -                 & $M_{f\sigma}(0,0)$ & $M_{f\sigma}(0,v)$ \\
															 $M_{f\sigma}(k,-v)$ & $M_{f\sigma}(k,0)$ & $M_{f\sigma}(k,v)$
                       \end{tabular} \right]
\end{eqnarray}
which corresponds to $N_{fmt}=(1+V)+K(2V+1)$ elements.

The discrete FMT approximation can be expressed in Cartesian coordinates without creating an image in polar coordinates:
\begin{eqnarray}
M^c_{f\sigma}(k,v) & = \frac{1}{2\pi} \sum\limits_{q=Q_{min}}^{Q_{max}} \sum\limits_{p=P_{min}}^{P_{max}} & f_c(p,q) (p+iq)^{-k}  \\
                   &                                                                                      &  \cdot (p^2+q^2)^{(k-2+\sigma-iv)/2} \nonumber
\end{eqnarray}
where $P_{min}$, $P_{max}$, $Q_{min}$, and $Q_{max}$ correspond to the set of coordinates describing the rectangle containing the object and where the position $(0,0)$ represents the gravity center of the image. 

Despite the fact that the magnitude of the complex moments can provide some rotation invariant features, the phase of the FMT includes substantial information related to the shape included in the image. To solve this issue, a set of invariant has been proposed to keep the information of the phase by normalizing with the first moments to compensate the change of scale and the rotation~\cite{ghorbel1994,derrode2001}. 

\begin{eqnarray}
I_{f\sigma}(k,v) & = & M^c_{f\sigma}(0,0)^{(-\sigma+iv)/\sigma} \cdot e^{ik arg(M^c_{f\sigma}(1,0))} \\
                 &  &  \cdot M^c_{f\sigma}(k,v) \nonumber
\end{eqnarray}
where the normalization in relation to the orientation is achieved through $M^c_{f\sigma}(0,0)$ and $M^c_{f\sigma}(1,0)$.

From the FMT, we can extract two sets of rotation invariant features:
\begin{eqnarray}
F_{fmt1} = |M^c_{f\sigma}(k,v)| ~~&~~ F_{fmt2} = I_{f\sigma}(k,v)
\end{eqnarray}
$F_{fmt1}$ has $N_{fmt}$ real values while $F_{fmt2}$ has $N_{fmt}$ complex values. 

\subsection{Normalization}

The different features in each set can be normalized by using transformations such as the z-score, i.e. by removing the mean and dividing by the standard deviation. Such a normalization can be beneficial in some techniques where there is no prior discriminant power on the different features. However, for the moment based techniques, it has been shown that the high order moments are more sensitive to the noise. Hence, a z-score normalization applied on all the moments can equalize the discriminant power of each moment and provide a lower performance when used with the Euclidean distance for the comparison of two objects.

\section{Datasets}
\label{sec:datasets}

\subsection{Artificial dataset}
\label{sec:artificial}

\subsubsection{Dataset description} 

\begin{figure*}[!t]
\centering
\begin{tabular}{ccccccccccc}
\includegraphics[width=0.12\figwidth]{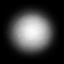} &
\includegraphics[width=0.12\figwidth]{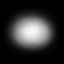} &
\includegraphics[width=0.12\figwidth]{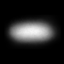} & 
\includegraphics[width=0.12\figwidth]{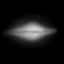} &
\includegraphics[width=0.12\figwidth]{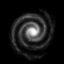} &
\includegraphics[width=0.12\figwidth]{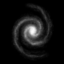} &
\includegraphics[width=0.12\figwidth]{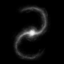} &
\includegraphics[width=0.12\figwidth]{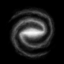} &
\includegraphics[width=0.12\figwidth]{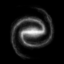} &
\includegraphics[width=0.12\figwidth]{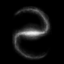} &
\includegraphics[width=0.12\figwidth]{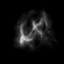} \\
E0 & E3 & E7 & S0 & Sa & Sb & Sc & SBa & SBb & SBc & I 
\end{tabular}
\caption{Templates for each type of galaxy (artificial images).}
\label{fig:template}
\end{figure*}

The artificial datasets are based on 11 types of galaxies: E0, E3, E7, S0, Sa, Sb, Sc, SBa, SBb, SBc, and I. Each type is represented by a graylevel image template of size 64 $\times$ 64. The original templates are depicted in Fig.~\ref{fig:template} with their respective class. Each image corresponds to a artificial representative example. For the evaluation of the different methods to extract rotation invariant features, we consider six main conditions: 1) the original template images, 2) the original images with speckle noise (corresponding to the superposition of stars), 3) the original images with Gaussian noise (corresponding to the removal of high frequency information), 4) the original images filtered with a Gaussian filter with a standard deviation of the Gaussian distribution $\sigma=2$; 5) same as condition 4 but with $\sigma=4$, and 6) the original images with the different types of noise (Speckle and Gaussian), filtered ($\sigma \in \{0.125, 1, 2, 4\}$. For each condition, we consider 12 different angles for each image ($\Delta\theta=\pi/6$). Therefore, there are 12 examples per class in conditions 1 to 5, and 144 examples for condition 6 (12 rotations, 4 types of Gaussian filtering, 3 types of noise (Speckle, Gaussian, no noise)). We denote each condition $i$ by its corresponding database $DBA_i$, e.g. $DBA_1$ for condition 1.

For the evaluation of these datasets, we consider three main cases. In the first case, we consider all the 11 classes: each class has 12 images. Here, it can be expected that for an example given as a test, it can retrieve the 11 other rotated images of the same image. This estimation aims at verifying the properties of rotation invariance for the different descriptors and the extent to which the calculation in the discrete case can lead to some ambiguities. In the second case, some of the initial classes are clustered together leading to 5 classes: S0, S (containing Sa, Sb, Sc), SB (containing SBa, SB, SBc), I, and E (containing E0, E3, E7). This analysis aims at estimating the potential confusions that may happen for instances within the same cluster. In the third case, we consider only 3 classes: S, SB, and E. For the first and third case, there is an equal number of images for each class, hence we report the precision to retrieve the total number of images from a class or cluster, and the average precision. For the second case, we report the precision at the rank equal to the minimum number of images in a class (12). For the second and third case, the goal of the evaluation is to determine if there exists a substantial difference between the images within a group and how these intragroup differences have an impact on the performance.

Before the feature extraction procedure, each image is normalized the following way: the image is centered on its gravity center, the maximum radius $R_{max}$ is extracted and the image is reduced to a square of size $2R_{max}$ with the gravity center being in the center of the image. The image is then resized using Bilinear interpolation to $62 \times 62$ and a border of 2 pixels is added around the image. It is worth nothing that for the detection of galaxy morphologies, there is no expected confusion between different classes contrary to the problem of multi oriented character recognition that leads to expected confusions, e.g., ('E','M'), ('Z','N'). 

\subsubsection{Parameters for the descriptors}

The number of descriptors for Hu and Flusser is fixed: 7 for $F_{Hu}$ and 11 for $F_{Flusser}$. For the estimation of $F_{zernike}$ and $F_{ring}$, we consider $N_{\rho}=10$ and $N_{\theta}=16$, giving a set of 12 and 40 features for $F_{zernike}$ and $F_{ring}$, respectively. For the FFT, we consider 32 points for the analysis in the Fourier domain, and $N_{\rho}=8$, which leads to 48 features for $F_{fft}$. For $F_{fmt1}$ and $F_{fmt2}$, we consider the parameters (k,v) such that $k=v=5$, $k=v=7$, and $k=v=9$. It provides 61, 113, and 181 features for k=5, 7, 9, respectively. For $F_{fmt2}$, we consider only the magnitude of the complex features to reduce the number of features.

\subsubsection{Classification and performance evaluation}

An image $x_i$ belongs to a class $C_i$, $1 \leq i \leq N_{class}$, with $N_{class}$ being the number of classes of images in the dataset. Each class $C_i$ has $N^i_{ex}$ examples. The classifier $E$ taking as an input the set of features representing the image $x$ returns a ranked list of $k$ examples $[y_1,\ldots,y_k]$, excluding $x$, sorted by increasing order of distances between $x$ and the elements in the list, i.e., $d(x,y_j) \leq d(x,y_{j+1})$, with $1 \leq j \leq k$.

The performance for each image $x_i$ belonging to the class $C_i$ is estimated by the average precision:
\begin{eqnarray}
\overline{P}(x_i) & = & \frac{1}{N^i_{ex}-1} \sum\limits_{k=1}^n P_k(x_i) \cdot rel_i(k)
\end{eqnarray}
where $N^i_{ex}-1$ represents the maximum number of relevant images to retrieve, i.e., the number of images belonging to the same class tested, minus the image tested itself. $rel_i(k)$ is defined as:
\begin{eqnarray}
rel_i(k) &  = & \left\{  \begin{tabular}{l} 1 if $y_k \in C_i$ \\ 0 otherwise\end{tabular}  \right. 
\end{eqnarray}

The precision is defined by:
\begin{eqnarray}
P_k(x_i) & = & \frac{1}{k}\sum\limits_{j=1}^k  \left\{  \begin{tabular}{l} 1 if $y_j \in C_i$ \\ 0 otherwise\end{tabular}  \right.
\end{eqnarray}
It counts the number of images in the list that belong to the class $C_i$. If the returned image at rank $j$ belongs to the class $C_i$, i.e., it is the same class as the evaluated example, then it is 1, 0 otherwise. When multiple classes are clustered and the number of examples in each cluster is different, we report the precision at rank $k_{min}$ where $k_{min}=min(N^i_{ex})$.

\subsection{Zoo Galaxy dataset}
\label{sec:zoo}

\subsubsection{Dataset description}

\begin{figure}[!t]
\centering
\includegraphics[width=0.95\figwidth]{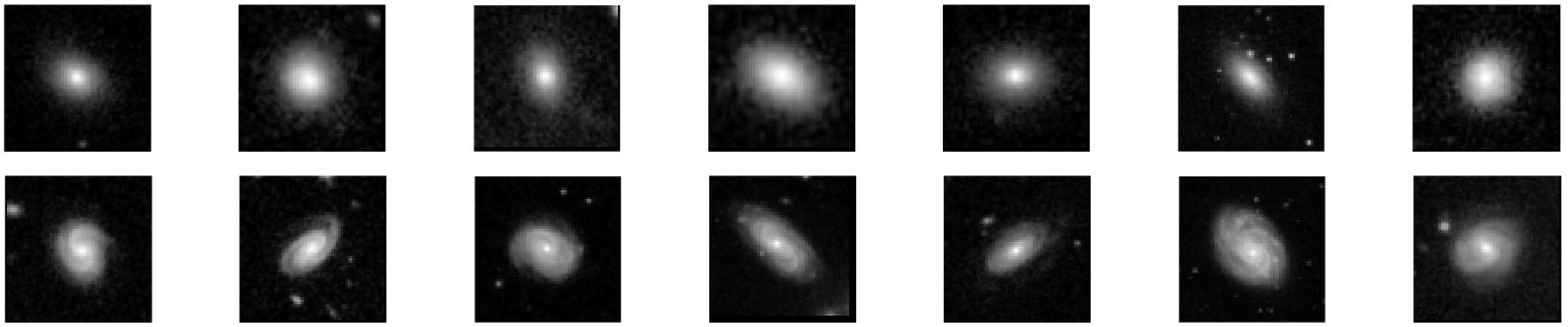}
\caption{Representative examples in graylevel. First row: elliptical galaxies; Second row: spiral galaxies.}
\label{fig:examples}
\end{figure}

In this problem, we consider the classification of real images of galaxies extracted from the Galaxy Zoo 2 (GZ2) dataset. All the details about the Zoo Galaxy dataset can be found in~\cite{willett2013}. In this dataset, a large number of images were given to participants who had to answer a series of questions (11 tasks and 37 answers) for each image. The groundtruth of the images corresponds to the score across all the participants, indicating a confidence score for all the different questions. We consider a subset of the whole dataset that contains 61578 color images. Among the questions, we select images that do not contain an ``odd'' element (i.e. the presence of a ring, a disturbed or irregular galaxy) with a confidence of at least 0.9. It limits the total of relevant images to 22481. In the present case, we limit the scope of the classification to elliptical versus spiral galaxies, which have been labeled with a confidence of at least 0.9, corresponding to 2517 elliptical and 2908 spiral galaxies. With the intersection of non-odd images, we finally obtain: 1545 elliptical and 884 spiral galaxies. Some representative images are depicted in Fig.~\ref{fig:examples}.

\subsubsection{Pre-processing}

Each image from the dataset has the size $424 \times 424$. First the images are transformed to graylevel and cropped, as we keep the center of the image with a size of $250 \times 250$. The images are then downsampled to $64 \times 64$. As the images are more noisy and may contain multiple elements such as stars. The image is binarized with the global thresholding Otsu method that separates pixels into foreground and background classes by minimizing intra-class intensity variance~\cite{otsu1979}. Then using morphological operators, we first apply the closing operation with a structuring element of size $5 \times 5$ (a square) followed by a dilation using a structuring element of size $13 \times 13$ (a disk) to compensate the problems related to the binarization with dark areas that still contain information. The binarization mask is then used to select the foreground of the original image. Finally, the image is centered on its gravity center $(m_{10}/m_{00},m_{01}/m_{00})$, the border is removed while keeping the gravity center in the center of the image, resized to $60 \times 60$, and a border of 2 pixels is added around the image, leading to an image of size $64 \times 64$.

\begin{figure}[!t]
\centering
	\subfigure[Elliptical galaxy.]{\includegraphics[width=0.95\figwidth]{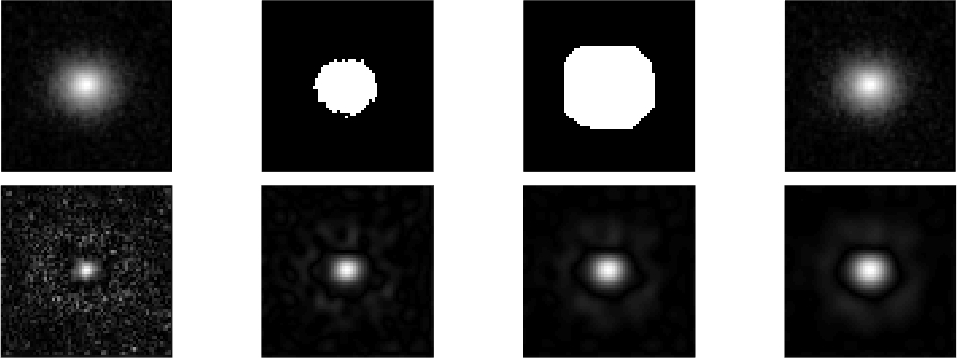}}
	\subfigure[Spiral galaxy.]{\includegraphics[width=0.95\figwidth]{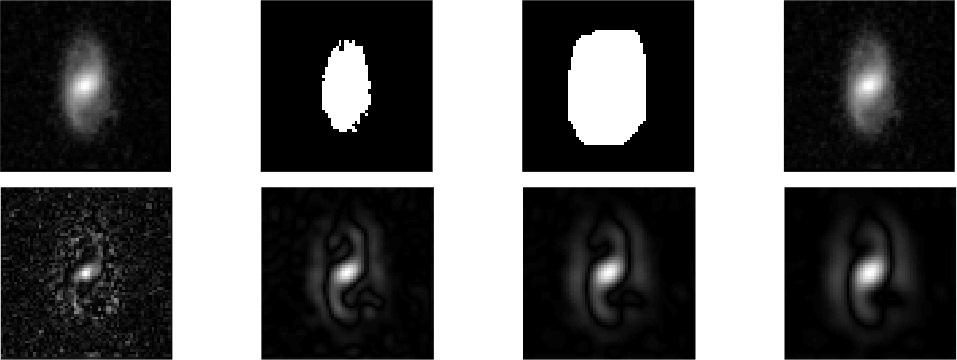}}
	\caption{Representative preprocessed steps. First row: original image, binarization mask, mask after morphological filtering, selected image. Second row: Laplacian pyramid.}
	\label{fig:preprocess}
\end{figure}

After normalizing the image, a Laplacian pyramid with 4 levels is applied on the image~\cite{burt1983}. The original image is convolved with a Gaussian kernel with a standard deviation of the Gaussian distribution set to 2. The Laplacian is then computed as the difference between the original image and the low pass filtered image. This process is repeated 4 times. This set of 4 images represents the input for the feature extraction part: the set of rotation invariant features is the concatenation of the rotation invariant feature sets applied to each of the 4 images. For each set of descriptors, the number of features is the same as for the artificial datasets, but multiplied by a factor 4, as there are multiple images given as an input. The different preprocessing steps are depicted in Fig.~\ref{fig:preprocess}.

\subsubsection{Performance evaluation}

For the binary classification of elliptical versus spiral galaxies, we consider 10-fold cross-validation procedure, with one block being used for the evaluation and the remaining 9 blocks for training. For each partition, the training dataset contains 1386 and 792 for the elliptical and spiral galaxies, respectively, while the test contains 154 and 88 examples. Given the unbalanced dataset in terms of number of examples per class, we report the area under the receiver operating characteristic curve (AUC)~\cite{fawcett2006}. In addition, we report the true positive rate (TPR), the false positive rate (FPR), the false negative rate (FNR), the true negative rate (TNR), and the f-score. All the scores represented in the subsequent sections correspond to the mean and standard deviation across the 10 partitions. They are defined by:

\begin{eqnarray}
\begin{tabular}{ll}
TPR=TP/P & FPR=FP/N \\
FNR=FN/P & TNR=TN/N \\
f-score = 2TP/(2TP+FP+FN) & 
\end{tabular}
\end{eqnarray}
where TP, FP, FN, TN corresponds to the true spirals, false spirals, false elliptical, and true elliptical, respectively. P and N corresponds to the total number of spiral and elliptical galaxies, respectively. 

\subsubsection{Classification}

For the binary classification, we consider the follow supervised learning techniques: Support Vector Machines (SVMs)~\cite{vapnik1998}, Bayesian Linear Discriminant Analysis (BLDA)~\cite{kay1992}, and stepwise Linear Discriminant Analysis (stepLDA)~\cite{draper1998}, and Extreme Learning Machine (ELM), a type of feedforward neural networks where the parameters of hidden units don't need to be tuned~\cite{huang2012}. These techniques provide a fast estimation of the classifiers with a minimum number of hyper-parameters~\cite{bishop2006}. The regular Linear Discriminant Analysis (LDA) is not considered as it requires some regularization technique due to the covariance matrices that was not always well conditioned on pilot tests with some descriptors. For the ELM, we consider 1000 hidden units and a sigmoid function as the activation for the random projections, which are normalized to obtain an an orthonormal basis of the random weights~\cite{huang2014}.

As pattern recognition systems for the classification of images using convolutional neural networks represent the state of the art~\cite{lecun2015}, we consider four architectures as a baseline to establish the relevance of rotation invariant descriptors. In each architecture, we use the Adam optimization algorithm~\cite{kingma2015}; the minibatch size is set to 64; the activation function for all the units is the rectified linear unit (ReLu)~\cite{glorot2011}; the maximum number of epochs is set to 100; the initial learning rate is set to 10e-4. The input layer is of size 64 $\times$ 64 while the output layer has 2 units (one for each class). The first architecture (A1) corresponds to a regular multilayer perceptron with only a single fully connected (FC) hidden layer of 100 units. The second architecture (A2) adds a convolutional layer with 5 feature maps, and filters of size 5 $\times$ 5 and stride of 2 in each direction, with no padding. The third architecture adds another convolutional layer (A3) with the same properties as the previous convolutional layer, with 25 feature maps. Finally the last architecture adds another convolutional layer (A4), with 125 feature maps, resulting in an architecture with 3 convolutional layers and a fully connected hidden layer. A2, A3, and A4 are CNNs. For each architecture, we consider two training conditions, with and without data augmentation based on the addition of rotated images ($\pm \pi/2$, $\pi$).

\section{Results}
\label{sec:results}

This section presents the performance for the artificial dataset results through an image retrieval angle, and for binary classification of elliptical versus spiral galaxies using real images.

\subsection{Artificial datasets}

The results for each condition and for each set of features are presented in Tables~\ref{table:c1},~\ref{table:c2},and~\ref{table:c3}. The first table corresponds to the original images from the chosen templates. These results highlight the problem related to the high dimensionality of the input feature set then used with the Euclidean distance, as there are shapes from different classes that are relatively similar. For $DBA_1$, $DBA_4$, and $DBA_5$, the best performance is obtained by Hu, Flusser moments, and the ring projection with FFT. These results show the low impact of Gaussian filtering on the images at the given scale. For $DBA_2$, with speckle noise, the best performance is obtained with Flusser moments with a precision of 98.83\%. However, with the addition of Gaussian noise, the best performance is achieved with the ring projection with FFT with a precision of 98.97\%. For $DBA_6$ when all the variations of the images are combined (rotation, noise, Gaussian filtering), the best precision reaches only 90.59\% with the the ring projection and FFT approach. It is worth noting the low performance of the Hu and Flusser approaches that provide only 58.77\% and 59.82\%, respectively, showing their low discriminant power when noise is added in the images. The best performance with the FMT is achieved with K=V=7 with 87.26\% by considering only the magnitude of the moments while the magnitude of the normalized moments provides only 75.49\%. In both cases, the choice of 7 for K and V provides a better performance than 5 and 9. 

When the classes are clustered, the Euclidean distance performs poorly with the descriptors based on Hu, Flusser, and Zernike with less than 80\% in average precision. The best method is $F_{fft}$ with an average precision of 83.36\%, followed by the $F_{fmt2}$ descriptors with k=v=9, with 82.30\%. These results highlight the robustness of the Fourier Mellin based descriptors when there exist a large intra-class variability. 

\begin{table*}[!tb]
\caption{Performance with the original multi oriented images ($DBA_1$) (\textbf{top}) and all the generated images ($DBA_6$) (\textbf{bottom}).}
\label{table:c1}
\centering
\begin{tabular}{|l|l|ccccc|} \hline
Dataset & Features & \multicolumn{2}{|c|}{11 classes} & 5 classes & \multicolumn{2}{|c|}{3 classes} \\
     &  & $P$ & $\overline{P}$ & $P$ & $P$ & $\overline{P}$ \\ \hline 										
& $F_{Hu}$ & $\mathbf{100.00\pm0.00}$ & $100.00\pm0.00$ & $100.00\pm0.00$ & $70.79\pm27.03$ & $79.47\pm19.15$ \\ 
& $F_{Flusser}$ & $\mathbf{100.00\pm0.00}$ & $100.00\pm0.00$ & $100.00\pm0.00$ & $73.33\pm23.79$ & $78.08\pm19.27$ \\ 
& $F_{Zernike}$ & $98.55\pm2.92$ & $99.06\pm2.03$ & $99.39\pm1.36$ & $71.59\pm25.98$ & $79.51\pm19.11$ \\ 
& $F_{ring}$ & $96.21\pm9.89$ & $98.32\pm4.88$ & $99.14\pm1.33$ & $61.80\pm3.33$ & $72.05\pm4.10$ \\ 
& $F_{fft}$ & $\mathbf{100.00\pm0.00}$ & $100.00\pm0.00$ & $100.00\pm0.00$ & $71.88\pm21.89$ & $82.40\pm14.02$ \\ 
$DBA_1$ & $F_{fmt1}$ (K=V=5) & $92.56\pm11.75$ & $95.05\pm8.15$ & $98.38\pm3.34$ & $71.16\pm16.50$ & $80.39\pm13.62$ \\ 
& $F_{fmt1}$ (K=V=7) & $94.42\pm9.01$ & $95.99\pm7.34$ & $98.18\pm3.79$ & $71.35\pm16.40$ & $80.43\pm13.58$ \\ 
& $F_{fmt1}$ (K=V=9) & $94.15\pm8.58$ & $95.84\pm7.27$ & $98.03\pm3.74$ & $71.01\pm17.18$ & $80.19\pm14.08$ \\ 
& $F_{fmt2}$ (K=V=5) & $77.20\pm18.01$ & $76.46\pm21.90$ & $87.07\pm13.75$ & $67.78\pm8.91$ & $74.54\pm6.11$ \\ 
& $F_{fmt2}$ (K=V=7) & $82.02\pm16.16$ & $80.66\pm20.88$ & $87.88\pm14.08$ & $71.69\pm13.99$ & $78.01\pm10.71$ \\ 
& $F_{fmt2}$ (K=V=9) & $89.88\pm11.88$ & $90.37\pm15.85$ & $91.52\pm15.61$ & $72.88\pm14.57$ & $83.95\pm10.70$ \\ 
\hline
& $F_{Hu}$ & $58.77\pm10.23$ & $66.60\pm12.12$ & $69.14\pm16.88$ & $67.04\pm27.52$ & $67.23\pm28.10$ \\ 
& $F_{Flusser}$ & $59.82\pm15.29$ & $65.05\pm16.93$ & $69.25\pm17.66$ & $65.97\pm27.42$ & $65.74\pm27.81$ \\
& $F_{Zernike}$ & $82.56\pm14.03$ & $86.10\pm16.68$ & $97.00\pm3.51$ & $71.51\pm25.45$ & $79.53\pm18.48$ \\  
& $F_{ring}$ & $75.53\pm10.42$ & $82.33\pm9.54$ & $88.44\pm7.66$ & $62.96\pm4.40$ & $71.67\pm10.07$ \\ 
& $F_{fft}$ & $\mathbf{90.59\pm13.42}$ & $94.14\pm9.09$ & $99.83\pm0.34$ & $72.52\pm22.78$ & $83.36\pm14.20$ \\ 
$DBA_6$ & $F_{fmt1}$ (K=V=5) & $86.98\pm13.66$ & $89.17\pm15.05$ & $97.77\pm3.26$ & $72.42\pm18.20$ & $81.70\pm13.90$ \\ 
& $F_{fmt1}$ (K=V=7) & $87.26\pm10.89$ & $91.20\pm9.71$ & $97.54\pm3.60$ & $72.49\pm18.21$ & $81.49\pm14.11$ \\ 
& $F_{fmt1}$ (K=V=9) & $83.70\pm12.40$ & $88.47\pm11.55$ & $96.87\pm4.42$ & $72.25\pm18.85$ & $80.65\pm15.00$ \\ 
& $F_{fmt2}$ (K=V=5) & $73.71\pm17.90$ & $73.62\pm20.00$ & $85.87\pm15.89$ & $67.78\pm7.52$ & $75.65\pm6.51$ \\ 
& $F_{fmt2}$ (K=V=7) & $75.49\pm15.33$ & $77.06\pm17.76$ & $86.22\pm16.29$ & $70.20\pm10.50$ & $77.64\pm8.92$ \\ 
& $F_{fmt2}$ (K=V=9) & $74.56\pm12.95$ & $77.59\pm15.77$ & $86.42\pm17.69$ & $74.52\pm15.02$ & $82.30\pm12.29$ \\ 
\hline
\end{tabular}
\end{table*}
							
\begin{table*}[!tb]
\caption{Performance with speckle noise ($DBA_2$) (top) and Gaussian noise ($DBA_3$) (bottom).}
\label{table:c2}
\centering
\begin{tabular}{|l|l|ccccc|} \hline
Dataset & Features & \multicolumn{2}{|c|}{11 classes} & 5 classes & \multicolumn{2}{|c|}{3 classes} \\
       & & $P$ & $\overline{P}$ & $P$ & $P$ & $\overline{P}$ \\ \hline											
& $F_{Hu}$ & $96.14\pm6.64$ & $96.55\pm6.54$ & $100.00\pm0.00$ & $71.11\pm26.60$ & $79.52\pm18.98$ \\ 
& $F_{Flusser}$ & $\mathbf{98.83\pm1.87}$ & $98.83\pm2.21$ & $100.00\pm0.00$ & $72.62\pm24.66$ & $78.45\pm19.17$ \\ 
& $F_{Zernike}$ & $88.71\pm18.33$ & $88.47\pm21.93$ & $98.89\pm2.35$ & $71.03\pm26.05$ & $79.04\pm19.00$ \\ 
& $F_{ring}$ & $90.36\pm10.07$ & $94.94\pm6.81$ & $96.92\pm3.27$ & $61.16\pm2.69$ & $73.63\pm6.00$ \\ 
& $F_{fft}$ & $97.80\pm3.90$ & $99.20\pm1.58$ & $99.95\pm0.11$ & $70.95\pm20.97$ & $81.77\pm13.92$ \\ 
$DBA_2$ & $F_{fmt1}$ (K=V=5) & $86.78\pm16.07$ & $85.68\pm19.36$ & $97.27\pm5.17$ & $71.11\pm16.89$ & $79.61\pm14.91$ \\ 
& $F_{fmt1}$ (K=V=7) & $89.88\pm11.59$ & $91.35\pm10.77$ & $97.07\pm5.61$ & $71.16\pm17.20$ & $79.24\pm15.41$ \\ 
& $F_{fmt1}$ (K=V=9) & $86.02\pm12.49$ & $89.00\pm11.66$ & $96.16\pm6.05$ & $70.85\pm18.14$ & $78.65\pm16.36$ \\ 
& $F_{fmt2}$ (K=V=5) & $74.72\pm18.12$ & $70.68\pm22.82$ & $86.11\pm17.60$ & $66.48\pm8.57$ & $73.68\pm5.14$ \\ 
& $F_{fmt2}$ (K=V=7) & $79.48\pm16.99$ & $78.65\pm20.98$ & $86.62\pm17.50$ & $69.50\pm12.71$ & $75.91\pm8.84$ \\ 
& $F_{fmt2}$ (K=V=9) & $81.06\pm14.38$ & $81.37\pm18.35$ & $87.27\pm21.36$ & $72.57\pm16.43$ & $80.85\pm11.23$ \\ 
\hline
& $F_{Hu}$ & $85.61\pm22.35$ & $83.82\pm27.07$ & $92.78\pm15.45$ & $70.19\pm27.85$ & $79.22\pm19.40$ \\ 
& $F_{Flusser}$ & $78.86\pm26.52$ & $75.20\pm31.55$ & $82.27\pm26.87$ & $70.90\pm26.89$ & $78.54\pm20.29$ \\
& $F_{Zernike}$ & $90.43\pm14.27$ & $90.19\pm17.08$ & $97.98\pm3.50$ & $71.48\pm25.89$ & $79.52\pm18.89$ \\  
& $F_{ring}$ & $82.23\pm12.07$ & $87.00\pm10.61$ & $96.52\pm3.26$ & $60.37\pm1.82$ & $70.93\pm5.90$ \\ 
& $F_{fft}$ & $\mathbf{98.97\pm2.12}$ & $99.66\pm0.74$ & $100.00\pm0.00$ & $72.43\pm22.66$ & $82.59\pm14.41$ \\ 
$DBA_3$ & $F_{fmt1}$ (K=V=5) & $89.33\pm12.90$ & $91.38\pm11.30$ & $97.63\pm3.59$ & $71.16\pm18.21$ & $79.84\pm14.96$ \\ 
& $F_{fmt1}$ (K=V=7) & $90.50\pm11.12$ & $92.84\pm9.48$ & $97.53\pm3.79$ & $71.22\pm18.32$ & $79.78\pm15.10$ \\ 
& $F_{fmt1}$ (K=V=9) & $88.71\pm12.43$ & $91.49\pm9.73$ & $96.77\pm4.65$ & $71.35\pm19.30$ & $79.10\pm16.08$ \\ 
& $F_{fmt2}$ (K=V=5) & $76.65\pm18.29$ & $75.71\pm22.09$ & $87.02\pm15.31$ & $66.96\pm6.72$ & $73.12\pm5.94$ \\ 
& $F_{fmt2}$ (K=V=7) & $80.17\pm16.75$ & $78.95\pm21.05$ & $87.93\pm15.34$ & $70.29\pm12.42$ & $75.90\pm10.63$ \\ 
& $F_{fmt2}$ (K=V=9) & $83.20\pm15.37$ & $84.51\pm17.52$ & $88.74\pm16.44$ & $72.88\pm15.52$ & $80.96\pm12.84$ \\ 
\hline
\end{tabular}
\end{table*}

\begin{table*}[!tb]
\caption{Performance with Gaussian filtering ($\sigma=2$) ($DBA_4$) (\textbf{top}) and ($\sigma=4$) ($DBA_5$) (\textbf{bottom}).}
\label{table:c3}
\centering
\begin{tabular}{|l|l|ccccc|} \hline
Dataset & Features & \multicolumn{2}{|c|}{11 classes} & 5 classes & \multicolumn{2}{|c|}{3 classes} \\
   &    & $P$ & $\overline{P}$ & $P$ & $P$ & $\overline{P}$ \\ \hline											
& $F_{Hu}$ & $\mathbf{100.00\pm0.00}$ & $100.00\pm0.00$ & $100.00\pm0.00$ & $70.79\pm27.03$ & $79.47\pm19.15$ \\ 
& $F_{Flusser}$ & $\mathbf{100.00\pm0.00}$ & $100.00\pm0.00$ & $100.00\pm0.00$ & $73.33\pm23.79$ & $78.65\pm18.95$ \\ 
& $F_{Zernike}$ & $98.76\pm2.45$ & $99.26\pm1.68$ & $99.85\pm0.34$ & $69.74\pm28.47$ & $79.17\pm19.56$ \\ 
& $F_{ring}$ & $96.56\pm8.25$ & $98.53\pm3.69$ & $100.00\pm0.00$ & $63.44\pm3.62$ & $75.01\pm7.81$ \\ 
& $F_{fft}$ & $\mathbf{100.00\pm0.00}$ & $100.00\pm0.00$ & $100.00\pm0.00$ & $71.06\pm21.85$ & $81.73\pm14.25$ \\ 
$DBA_4$ & $F_{fmt1}$ (K=V=5) & $95.94\pm9.17$ & $97.68\pm5.81$ & $99.49\pm1.13$ & $71.96\pm16.80$ & $81.08\pm13.27$ \\ 
& $F_{fmt1}$ (K=V=7) & $97.59\pm5.22$ & $98.98\pm2.20$ & $99.34\pm1.33$ & $72.06\pm16.61$ & $81.38\pm12.87$ \\ 
& $F_{fmt1}$ (K=V=9) & $97.31\pm5.39$ & $98.72\pm2.70$ & $99.19\pm1.54$ & $72.14\pm17.46$ & $81.14\pm13.49$ \\ 
& $F_{fmt2}$ (K=V=5) & $78.17\pm16.61$ & $76.78\pm21.24$ & $86.67\pm13.67$ & $64.15\pm7.47$ & $72.39\pm5.03$ \\ 
& $F_{fmt2}$ (K=V=7) & $82.30\pm15.90$ & $81.53\pm20.61$ & $87.02\pm13.74$ & $67.12\pm10.70$ & $74.78\pm7.18$ \\ 
& $F_{fmt2}$ (K=V=9) & $89.05\pm12.74$ & $90.15\pm15.48$ & $90.30\pm15.22$ & $72.75\pm13.96$ & $83.42\pm10.83$ \\ 
\hline
& $F_{Hu}$ & $\mathbf{100.00\pm0.00}$ & $100.00\pm0.00$ & $100.00\pm0.00$ & $70.79\pm27.03$ & $79.47\pm19.15$ \\ 
& $F_{Flusser}$ & $\mathbf{100.00\pm0.00}$ & $100.00\pm0.00$ & $100.00\pm0.00$ & $73.33\pm23.79$ & $78.65\pm18.95$ \\ 
& $F_{Zernike}$ & $98.90\pm2.43$ & $99.30\pm1.66$ & $99.95\pm0.11$ & $69.63\pm28.61$ & $79.19\pm19.54$ \\ 
& $F_{ring}$ & $96.76\pm8.16$ & $98.53\pm3.68$ & $100.00\pm0.00$ & $63.78\pm2.93$ & $75.21\pm7.48$ \\ 
& $F_{fft}$ & $\mathbf{100.00\pm0.00}$ & $100.00\pm0.00$ & $100.00\pm0.00$ & $70.85\pm21.96$ & $81.64\pm14.28$ \\ 
$DBA_5$ & $F_{fmt1}$ (K=V=5) & $96.01\pm9.14$ & $97.73\pm5.70$ & $99.55\pm1.02$ & $72.12\pm16.89$ & $81.11\pm13.31$ \\ 
& $F_{fmt1}$ (K=V=7) & $97.80\pm5.08$ & $99.05\pm2.10$ & $99.49\pm1.13$ & $72.30\pm16.68$ & $81.46\pm12.87$ \\ 
& $F_{fmt1}$ (K=V=9) & $97.52\pm5.21$ & $98.82\pm2.48$ & $99.34\pm1.33$ & $72.22\pm17.44$ & $81.24\pm13.46$ \\ 
& $F_{fmt2}$ (K=V=5) & $78.24\pm16.52$ & $77.05\pm21.07$ & $86.67\pm13.67$ & $63.52\pm7.04$ & $72.18\pm4.87$ \\ 
& $F_{fmt2}$(K=V=7) & $82.44\pm15.86$ & $81.81\pm20.58$ & $87.07\pm13.75$ & $66.75\pm10.49$ & $74.52\pm6.84$ \\ 
& $F_{fmt2}$ (K=V=9) & $89.26\pm12.60$ & $90.31\pm15.42$ & $90.30\pm15.22$ & $72.67\pm14.04$ & $83.32\pm10.90$ \\ 
\hline
\end{tabular}
\end{table*}	
												
\subsection{Galaxy Zoo}

The performance for the different type of features are detailed in Tables~\ref{table:gz01} and~\ref{table:gz02} with values in \%. The best performance for each table is given in bold. The best performance is obtained with the combination of the ring projection, and the stepLDA classifier with an average AUC=99.54. The second best technique is the ring projection with FFT in each ring combined with BLDA. The performance of ELM with only AUC=97.09 suggests there is no need to add an extra level in the architecture and simple linear classifiers are enough given the provided input features. The level of performance across the different descriptors is relatively similar than the artificial dataset ($DBA_6$): Hu and Flusser based moments do not provide the best descriptors. 

\begin{table*}[!tb]
\caption{Performance on the Galaxy Zoo images with SVM (\textbf{top}) and ELM (\textbf{bottom}).}
\label{table:gz01}
\centering
\begin{tabular}{|l|l|cc|cccc|} \hline
Classifier & Features & AUC & f-score & TPR  & FPR & FNR & TNR \\ \hline
& $F_{Hu}$ & $88.13\pm2.53$ & $76.95\pm3.38$ & $76.36\pm6.14$ & $12.60\pm4.59$ & $23.64\pm6.14$ & $87.40\pm4.59$ \\ 
& $F_{Flusser}$ & $91.95\pm2.75$ & $82.59\pm3.42$ & $85.68\pm7.06$ & $12.34\pm3.82$ & $14.32\pm7.06$ & $87.66\pm3.82$ \\
& $F_{Zernike}$ & $98.67\pm1.05$ & $93.67\pm3.14$ & $95.45\pm3.48$ & $4.81\pm2.82$ & $4.55\pm3.48$ & $95.19\pm2.82$ \\  
& $F_{ring}$ & $\mathbf{99.48\pm0.53}$ & $96.85\pm1.50$ & $97.50\pm1.82$ & $2.21\pm1.54$ & $2.50\pm1.82$ & $97.79\pm1.54$ \\ 
& $F_{fft}$ & $99.12\pm0.83$ & $95.74\pm1.75$ & $95.57\pm1.79$ & $2.34\pm1.27$ & $4.43\pm1.79$ & $97.66\pm1.27$ \\ 
SVM & $F_{fmt1}$ (K=V=5) & $97.84\pm0.78$ & $92.80\pm2.09$ & $93.52\pm3.01$ & $4.61\pm2.59$ & $6.48\pm3.01$ & $95.39\pm2.59$ \\ 
& $F_{fmt1}$ (K=V=7) & $98.14\pm0.60$ & $93.19\pm1.48$ & $93.30\pm2.24$ & $3.96\pm1.49$ & $6.70\pm2.24$ & $96.04\pm1.49$ \\ 
& $F_{fmt1}$ (K=V=9) & $98.13\pm0.91$ & $93.60\pm1.46$ & $93.07\pm3.39$ & $3.31\pm2.02$ & $6.93\pm3.39$ & $96.69\pm2.02$ \\ 
& $F_{fmt2}$ (K=V=5) & $96.68\pm0.86$ & $90.65\pm1.60$ & $92.16\pm3.19$ & $6.36\pm1.50$ & $7.84\pm3.19$ & $93.64\pm1.50$ \\ 
& $F_{fmt2}$ (K=V=7) & $96.77\pm1.49$ & $91.71\pm2.25$ & $93.18\pm2.16$ & $5.78\pm2.75$ & $6.82\pm2.16$ & $94.22\pm2.75$ \\ 
& $F_{fmt2}$ (K=V=9) & $96.95\pm1.89$ & $92.69\pm1.70$ & $93.75\pm2.89$ & $4.87\pm1.52$ & $6.25\pm2.89$ & $95.13\pm1.52$ \\  
\hline
& $F_{Hu}$ & $54.18\pm3.36$ & $40.76\pm2.04$ & $42.27\pm2.02$ & $57.73\pm2.02$ & $37.27\pm2.95$ & $62.73\pm2.95$ \\ 
& $F_{Flusser}$ & $53.67\pm3.72$ & $40.48\pm5.19$ & $41.93\pm6.85$ & $58.07\pm6.85$ & $36.69\pm3.01$ & $63.31\pm3.01$ \\ 
& $F_{Zernike}$ & $95.09\pm1.51$ & $90.91\pm1.86$ & $89.32\pm2.93$ & $10.68\pm2.93$ & $4.09\pm1.67$ & $95.91\pm1.67$ \\ 
& $F_{ring}$ & $85.58\pm3.69$ & $80.57\pm3.44$ & $81.25\pm3.78$ & $18.75\pm3.78$ & $11.69\pm2.75$ & $88.31\pm2.75$ \\ 
& $F_{fft}$ & $96.93\pm2.21$ & $93.70\pm2.07$ & $92.39\pm3.13$ & $7.61\pm3.13$ & $2.73\pm1.08$ & $97.27\pm1.08$ \\ 
ELM & $F_{fmt1}$ (K=V=5) & $96.04\pm1.57$ & $90.55\pm1.97$ & $89.32\pm2.17$ & $10.68\pm2.17$ & $4.55\pm1.42$ & $95.45\pm1.42$ \\ 
& $F_{fmt1}$ (K=V=7) & $\mathbf{97.09\pm1.17}$ & $91.05\pm1.29$ & $88.07\pm2.70$ & $11.93\pm2.70$ & $3.05\pm0.92$ & $96.95\pm0.92$ \\ 
& $F_{fmt1}$ (K=V=9) & $96.19\pm1.04$ & $91.10\pm1.47$ & $89.20\pm3.18$ & $10.80\pm3.18$ & $3.77\pm1.08$ & $96.23\pm1.08$ \\ 
& $F_{fmt2}$ (K=V=5) & $94.50\pm2.13$ & $89.61\pm1.69$ & $89.20\pm2.45$ & $10.80\pm2.45$ & $5.65\pm1.72$ & $94.35\pm1.72$ \\ 
& $F_{fmt2}$ (K=V=7) & $95.31\pm1.28$ & $90.19\pm1.87$ & $88.86\pm3.00$ & $11.14\pm3.00$ & $4.68\pm1.90$ & $95.32\pm1.90$ \\ 
& $F_{fmt2}$ (K=V=9) & $95.71\pm1.57$ & $90.95\pm2.23$ & $90.00\pm3.80$ & $10.00\pm3.80$ & $4.48\pm1.31$ & $95.52\pm1.31$ \\ 
\hline
\end{tabular}
\end{table*}

\begin{table*}[!tb]
\caption{Performance on the Galaxy Zoo images with BLDA (\textbf{top}) and stepLDA (\textbf{bottom}).}
\label{table:gz02}
\centering
\begin{tabular}{|l|l|cc|cccc|} \hline
Classifier & Features & AUC & f-score & TPR  & FPR & FNR & TNR \\ \hline
& $F_{Hu}$ & $86.00\pm3.40$ & $74.04\pm3.66$ & $77.16\pm6.13$ & $17.99\pm6.79$ & $22.84\pm6.13$ & $82.01\pm6.79$ \\ 
& $F_{Flusser}$ & $82.53\pm18.14$ & $75.25\pm9.61$ & $86.36\pm7.77$ & $27.60\pm23.79$ & $13.64\pm7.77$ & $72.40\pm23.79$ \\
& $F_{Zernike}$ & $98.92\pm0.49$ & $94.35\pm2.02$ & $96.14\pm3.14$ & $4.35\pm1.33$ & $3.86\pm3.14$ & $95.65\pm1.33$ \\  
& $F_{ring}$ & $93.07\pm4.79$ & $83.77\pm8.71$ & $86.36\pm7.74$ & $11.56\pm6.89$ & $13.64\pm7.74$ & $88.44\pm6.89$ \\ 
& $F_{fft}$ & $\mathbf{99.50\pm0.58}$ & $96.70\pm0.99$ & $96.70\pm2.24$ & $1.88\pm0.84$ & $3.30\pm2.24$ & $98.12\pm0.84$ \\ 
BLDA & $F_{fmt1}$ (K=V=5) & $98.99\pm0.53$ & $94.67\pm1.59$ & $93.98\pm2.54$ & $2.60\pm1.36$ & $6.02\pm2.54$ & $97.40\pm1.36$ \\ 
& $F_{fmt1}$ (K=V=7) & $99.06\pm0.43$ & $95.34\pm2.03$ & $95.23\pm2.32$ & $2.60\pm1.36$ & $4.77\pm2.32$ & $97.40\pm1.36$ \\ 
& $F_{fmt1}$ (K=V=9) & $99.08\pm0.46$ & $95.58\pm1.78$ & $95.68\pm2.37$ & $2.60\pm1.74$ & $4.32\pm2.37$ & $97.40\pm1.74$ \\ 
& $F_{fmt2}$ (K=V=5) & $98.06\pm0.78$ & $92.82\pm1.71$ & $94.66\pm2.10$ & $5.32\pm1.53$ & $5.34\pm2.10$ & $94.68\pm1.53$ \\ 
& $F_{fmt2}$ (K=V=7) & $98.35\pm0.73$ & $93.45\pm1.99$ & $93.30\pm2.71$ & $3.64\pm1.17$ & $6.70\pm2.71$ & $96.36\pm1.17$ \\ 
& $F_{fmt2}$ (K=V=9) & $98.46\pm0.72$ & $94.41\pm1.65$ & $94.77\pm2.75$ & $3.44\pm2.10$ & $5.23\pm2.75$ & $96.56\pm2.10$ \\ 
\hline
& $F_{Hu}$ & $88.67\pm2.60$ & $75.73\pm4.50$ & $71.36\pm5.92$ & $28.64\pm5.92$ & $9.61\pm1.26$ & $90.39\pm1.26$ \\ 
& $F_{Flusser}$ & $92.64\pm1.94$ & $79.31\pm4.34$ & $74.43\pm4.83$ & $25.57\pm4.83$ & $7.53\pm2.40$ & $92.47\pm2.40$ \\
& $F_{Zernike}$ & $98.82\pm0.52$ & $91.85\pm2.13$ & $89.32\pm3.38$ & $10.68\pm3.38$ & $2.92\pm1.24$ & $97.08\pm1.24$ \\  
& $F_{ring}$ & $\mathbf{99.54\pm0.51}$ & $95.64\pm1.81$ & $93.86\pm2.93$ & $6.14\pm2.93$ & $1.36\pm0.68$ & $98.64\pm0.68$ \\ 
& $F_{fft}$ & $99.36\pm0.60$ & $95.51\pm0.93$ & $94.32\pm1.61$ & $5.68\pm1.61$ & $1.82\pm0.39$ & $98.18\pm0.39$ \\ 
stepLDA & $F_{fmt1}$ (K=V=5) & $98.62\pm0.79$ & $93.99\pm1.38$ & $93.30\pm1.56$ & $6.70\pm1.56$ & $2.99\pm1.17$ & $97.01\pm1.17$ \\ 
& $F_{fmt1}$ (K=V=7) & $98.72\pm0.48$ & $93.12\pm2.32$ & $91.70\pm3.26$ & $8.30\pm3.26$ & $2.99\pm1.67$ & $97.01\pm1.67$ \\ 
& $F_{fmt1}$ (K=V=9) & $98.34\pm0.67$ & $93.26\pm1.82$ & $91.25\pm2.49$ & $8.75\pm2.49$ & $2.53\pm1.41$ & $97.47\pm1.41$ \\ 
& $F_{fmt2}$ (K=V=5) & $97.89\pm0.94$ & $91.19\pm2.49$ & $90.11\pm2.92$ & $9.89\pm2.92$ & $4.29\pm1.37$ & $95.71\pm1.37$ \\ 
& $F_{fmt2}$ (K=V=7) & $97.96\pm0.86$ & $91.94\pm1.99$ & $90.57\pm2.22$ & $9.43\pm2.22$ & $3.70\pm1.88$ & $96.30\pm1.88$ \\ 
& $F_{fmt2}$ (K=V=9) & $97.21\pm1.36$ & $92.32\pm2.57$ & $90.91\pm3.37$ & $9.09\pm3.37$ & $3.44\pm1.84$ & $96.56\pm1.84$ \\ 
\hline
\end{tabular}
\end{table*}	

The results for the artificial neural networks are presented in Table~\ref{table:gz03}. The performance with only a single hidden layer reaches an AUC of 90.36\% whereas the performance increases when convolutional layers are added. The best performance is obtained with the architecture using 3 convolutional layers, with an AUC of 96.81\%. This performance is inferior to what could be obtained with the best rotation invariant descriptors, e.g., $F_{ring}$, $F_{fft}$, but it remains superior to $F_{Hu}$ and $F_{Flusser}$. These results indicate that the variability that exists within the images can be captured by the neural networks, and that it is better modeled through the use of convolutional layers. With the addition of rotated images in the training dataset, the performance substantially improved in all the architectures, reaching an AUC of 99.04\% with A4, which is very close to the best performance obtained with rotation invariant descriptors. These results suggest that despite having images with various orientations in the training database, it is necessary to enrich the training database to improve the performance.

\begin{table*}[!tb]
\caption{Performance on the Galaxy Zoo images with artificial neural networks and images as inputs.}
\label{table:gz03}
\centering
\begin{tabular}{|l|l|cc|cccc|} \hline
Architecture & Data augmentation & AUC & f-score & TPR  & FPR & FNR & TNR \\ \hline
A1 & No  & $90.36\pm1.67$ & 86.87 & 92.40 & 35.57 & 7.60 & 64.43 \\
A2 & No  & $94.28\pm1.08$ & 90.38 & 91.81 & 19.89 & 8.18 & 80.11 \\ 
A3 & No  & $95.40\pm0.81$ & 91.76 & 92.21 & 15.34 & 7.79 & 84.66 \\ 
A4 & No  & $96.81\pm0.71$ & 92.89 & 91.62 & 9.89 & 8.38 & 90.11 \\ \hline
A1 & Yes & $93.69\pm1.81$ & 90.16 & 92.27 & 21.70 & 7.73 & 78.29 \\
A2 & Yes & $96.75\pm0.83$ & 93.78 & 94.54 & 12.39 & 5.45 & 87.61 \\ 
A3 & Yes & $98.37\pm0.40$ & 95.54 & 96.10 & 8.86 & 3.89 & 91.14 \\ 
A4 & Yes & $99.04\pm0.45$ & 96.47 & 95.97 & 5.23 & 4.03 & 94.77 \\
\hline
\end{tabular}
\end{table*}	

\subsection{Relationships between classification and human confidence}

\begin{figure}[!t]
\centering
	\begin{tabular}{c}
  \subfigure[Distribution]{\includegraphics[width=0.8\figwidth]{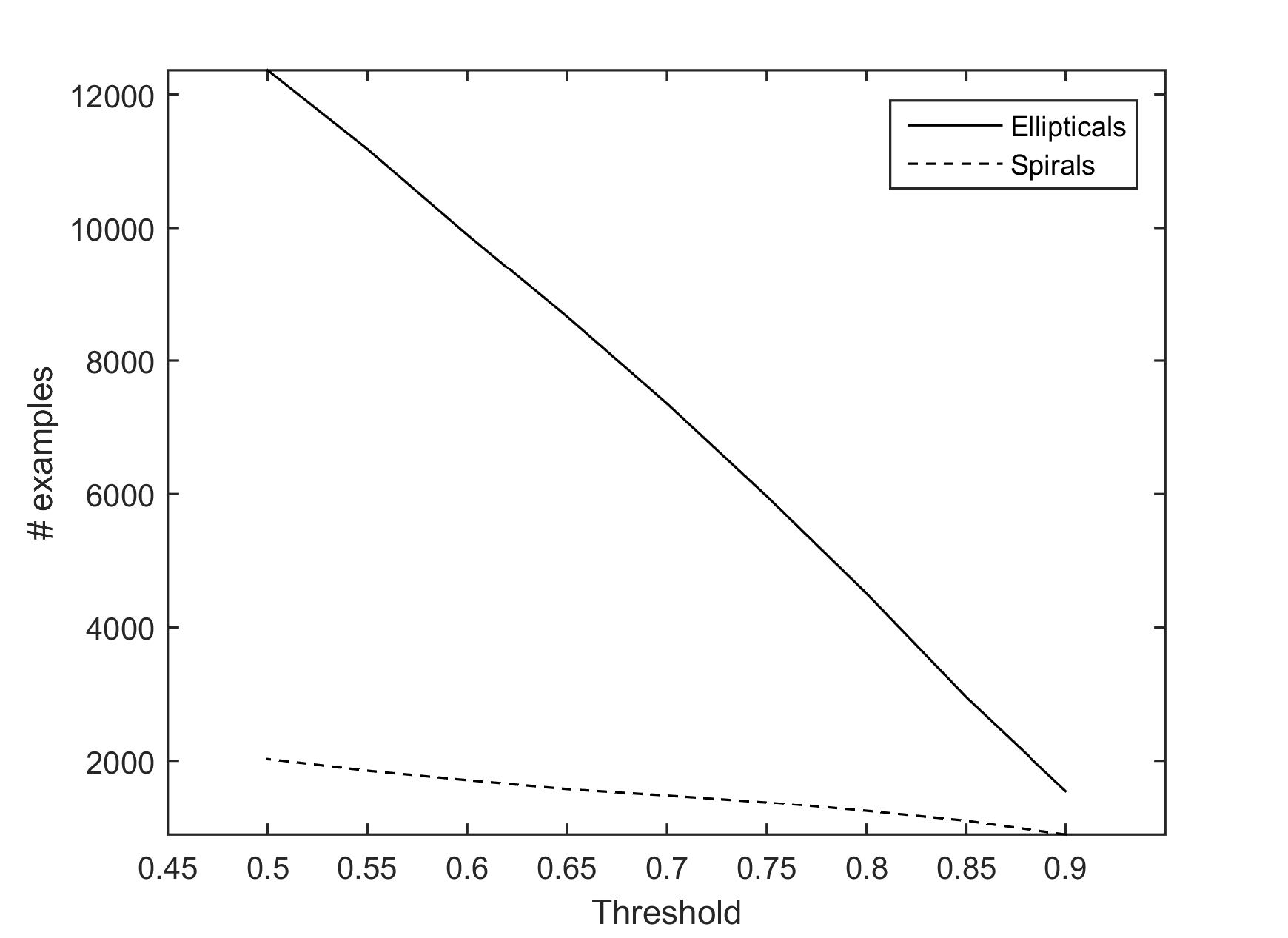}} \\
	\subfigure[AUC]{\includegraphics[width=0.8\figwidth]{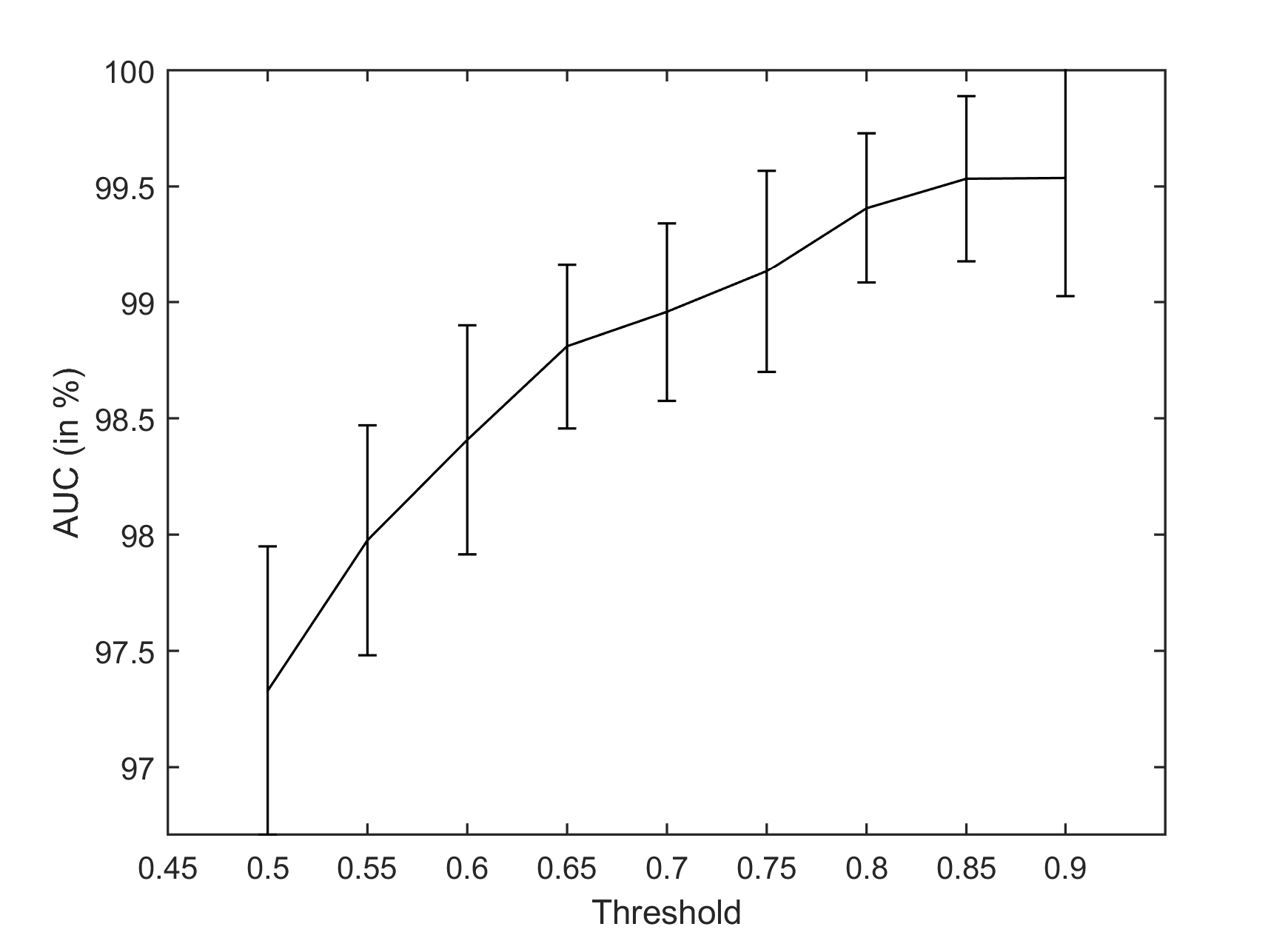}} \\
	\subfigure[f-score]{\includegraphics[width=0.8\figwidth]{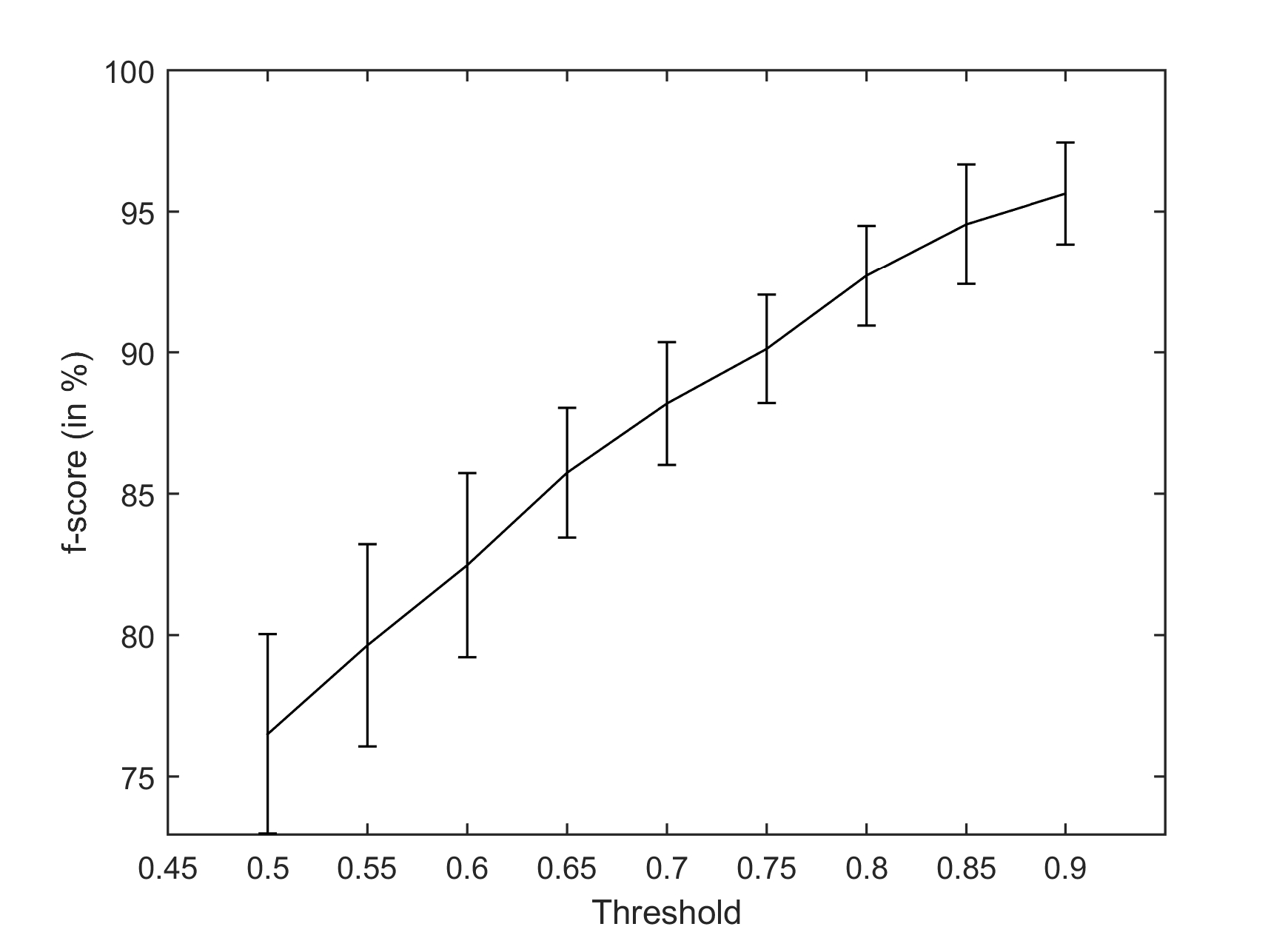}}
	\end{tabular}
	\caption{Performance in relation to the selected confidence for the selection of the images in the Galaxy Zoo 2 dataset.}
	\label{fig:confidence}
\end{figure}

The selected images in the Galaxy Zoo 2 dataset included only images where the confidence score was above 90\%. Fig.~\ref{fig:confidence} represents the relationship between the global confidence and the classification score obtained with $F_{ring}$ and stepLDA. The distribution of the number of examples in relation to the chosen threshold highlights the substantial increase of the number of examples in the dataset. As expected, the performance of the classifier decreases as a function of the chosen threshold due to the addition of more difficult images. The AUC remains nevertheless above 97\% for all the chosen thresholds.

\section{Discussion} 
\label{sec:discussion}

A comprehensive description of the main families of rotation invariant features descriptors has been described in this paper, including moments of Hu, Flusser, Zernike, and Fourier-Mellin. The ring projection technique has been defined using both 1D moments and the Fourier transform through the analysis of log decomposition of the bandpowers. These techniques have been evaluated and compared on different datasets relative to the classification of galaxy morphologies, highlighting the strengths and pitfalls of these approaches. Using these techniques, a complete framework has been proposed for fast classification with low level features based on a Laplacian pyramid and pre-processing based on morphological filtering. These techniques have been compared with convolutional neural networks, which represent the state of the art for image classification tasks.

The results obtained with the artificial dataset confirms the interest of all the different techniques as descriptors when the images are clean. However, there is a tradeoff between the dimensionality of the input vectors when used with distances, and the level of noise in the images. The results confirm the issues with Hu and Flusser moments in noisy images. The results obtained with the real Galaxy Zoo dataset highlight the possibility to discriminate with a high performance images of galaxies. Following the same pattern of performance as with the artificial images, the Hu and Flusser moments provide the worst accuracy, albeit above 80\%. There was not a significant difference of performance across the binary classifiers as all the results remain in the range of $99.50\pm0.50$\%. The worst classifier was ELM with a maximum accuracy of 97.09\%. These results suggest that the use of morphological filtering for denoising images followed by Laplacian pyramid as input features and the use of rotation invariant descriptors give state of the art results that can be exploited by astronomers for galaxy morphology classification. These results are higher than the results obtained with the convolutional neural networks that had 97\%, then data augmentation was not present. However, the results are relatively similar (above 99\%) when data augmentation is used in the training dataset, confirming that CNNs are robust classifiers in such an application despite the large variability that exists across examples in relation to their different orientations. In both cases, the image processing pipeline on one hand, and the neural network architecture and its parameters on the other hand, are designed in relation to type of input image to classify.

While this paper is dealing with the effect of the rotation, the scale and the normalization of the images has a key impact on the choice of the method. Before the extraction of descriptors that are invariant to the rotation, the first problem to solve is the invariance to the translation, which is typically dealt with the normalization to the gravity center of the image. In noisy images, this step can have a significant impact on the results. In addition, the normalization of the scale of the image is a key step for the methods that rely directly on the radius such as the ring projection. The difficulty related to the estimation of the maximum radius of the image can prevent the use of the ring projection and the direct FFT approach on the polar representation of the images due to the variability of the optimal radius across images.

The creation of a pattern recognition system for object classification is difficult as it requires multiple stages, from feature selection, reduction, and/or extraction to the classification stage. In relation to a given problem, features can be extracted analytically or through deep learning approaches. The latter approach has obtained great performance compared to the former in computer vision. Yet, the use of descriptors that are invariant to translation and rotation allows to have an estimation of the differences between the different classes of objects without requiring a large number of labeled examples. CNNs are relevant classifiers applied to images but it remains difficult to understand what happens within the last hidden layers for extracting high level features. Furthermore, CNNs are able to absorb a large number of variations (translation, rotation, scaling,...) through data augmentation, which requires to know how to enrich the initial training database. CNNs have been recently used for classifying radio galaxies~\cite{aniyan2017} and for the Zoo Galaxy projects~\cite{dieleman2015}. Thanks to the low level features that can be extracted by CNN models, it would be possible to replace the Laplacian pyramid that has been used in this work by feature maps obtained from CNNs. Transfer learning through the use of CNNs based features may provide a better feature set to apply translation, rotation, and scale invariant descriptors. Furthermore, the research in CNN architectures should enable the use of functions connecting layers to extract directly rotation invariant features without introducing as an input all the different possible orientations. Other novel approaches such as capsule networks provide promising results for galaxy morphology classification~\cite{katebi2019}.

Most of the frameworks for the classification of galaxies remain supervised. Data labeled by regular citizen can be used training machine learning systems and the performance depends on the quality of the data, which can be improved by using images that have a high degree of agreement across participants~\cite{kuminski2014}. The involvement of the participants is currently separated from the machine learning part, preventing the use of active learning techniques that can combine dynamically manual labeling and machine learning in order to minimize the amount of manual work while keeping a high reliability in the decisions~\cite{cecotti2016}. The definition of efficient descriptors for the classification is one step toward the definition of efficient distances that can be considered for graph based semi-supervised learning that can estimate the characteristics of galaxies using labeled images, unlabeled images, and query human participants in an active learning setting.

\section{Conclusion} 
\label{sec:conclusion}

In this paper, we have provided a comprehensive description of the main families of rotation invariant descriptors that can be considered for the analysis of galaxy morphologies. Six main techniques have been presented and compared in different datasets, with artificial images in the context of content retrieval, and for the classification of images using real images of galaxies. The results have been compared with convolutional neural networks and highlighted the low difference in terms of performance between the descriptors. The ring based methods using 1D moments or the Fast Fourier Transform have provided the best performance then used with linear binary classifiers. With an AUC reaching 99.54\%, these encouraging results suggest the possibility to determine finer galaxy morphological characteristics by using the proposed image processing framework combining a Laplacian pyramid, rotation translation invariant descriptors, and state of the art binary classifiers. Future work will deal with the effect of the errors related to the estimation of the gravity center and the presence of external elements within the shape that can disturb its analysis.

\bibliographystyle{IEEEtran}      
\bibliography{references}   

\begin{thebibliography}{10}
\providecommand{\url}[1]{#1}
\csname url@samestyle\endcsname
\providecommand{\newblock}{\relax}
\providecommand{\bibinfo}[2]{#2}
\providecommand{\BIBentrySTDinterwordspacing}{\spaceskip=0pt\relax}
\providecommand{\BIBentryALTinterwordstretchfactor}{4}
\providecommand{\BIBentryALTinterwordspacing}{\spaceskip=\fontdimen2\font plus
\BIBentryALTinterwordstretchfactor\fontdimen3\font minus
  \fontdimen4\font\relax}
\providecommand{\BIBforeignlanguage}[2]{{%
\expandafter\ifx\csname l@#1\endcsname\relax
\typeout{** WARNING: IEEEtran.bst: No hyphenation pattern has been}%
\typeout{** loaded for the language `#1'. Using the pattern for}%
\typeout{** the default language instead.}%
\else
\language=\csname l@#1\endcsname
\fi
#2}}
\providecommand{\BIBdecl}{\relax}
\BIBdecl

\bibitem{ball2010}
N.~M. Ball and R.~J. Brunner, ``Data mining and machine learning in
  astronomy,'' \emph{International Journal of Modern Physics D}, vol.~19,
  no.~7, pp. 1049--1106, 2010.

\bibitem{2011MNRAS.410..166L}
C.~{Lintott}, K.~{Schawinski}, S.~{Bamford}, A.~{Slosar}, K.~{Land},
  D.~{Thomas}, E.~{Edmondson}, K.~{Masters}, R.~C. {Nichol}, M.~J. {Raddick},
  A.~{Szalay}, D.~{Andreescu}, P.~{Murray}, and J.~{Vandenberg}, ``{Galaxy Zoo
  1: data release of morphological classifications for nearly 900 000
  galaxies},'' \emph{Monthly Notices of the Royal Astronomical Society}, vol.
  410, pp. 166--178, Jan. 2011.

\bibitem{willett2013}
K.~W. Willett, C.~J. Lintott, S.~P. Bamford, K.~L. Masters, K.~R. Simmons,
  Brooke D.~Casteels, E.~M. Edmondson, L.~F. Fortson, S.~Kaviraj, W.~C. Keel,
  T.~Melvin, R.~C. Nichol, M.~J. Raddick, K.~Schawinski, R.~J. Simpson, R.~A.
  Skibba, A.~M. Smith, and D.~Thomas, ``Galaxy zoo 2: detailed morphological
  classifications for 304,122 galaxies from the sloan digital sky survey,''
  \emph{Monthly Notices of the Royal Astronomical Society}, vol. 435, p.
  1–29, 2013.

\bibitem{shamir2009}
L.~Shamir, ``Automatic morphological classification of galaxy images,''
  \emph{Monthly Notices of the Royal Astronomical Society}, vol. 399, no.~3,
  pp. 1367--1372, 2009.

\bibitem{shamir2016}
L.~Shamir, D.~Diamond, and J.~Wallin, ``Leveraging pattern recognition
  consistency estimation for crowdsourcing data analysis,'' \emph{IEEE Trans.
  on Human-Machine Systems}, vol.~46, pp. 474--480, June 2016.

\bibitem{hubble1926}
E.~P. Hubble, ``Extra-galactic nebulae,'' \emph{Astrophysical Journal},
  vol.~64, pp. 321--369, 1926.

\bibitem{buta2001}
R.~J. Buta, ``Galaxies: Classification,'' in \emph{Encyclopedia of Astronomy
  and Astrophysics}, P.~Murdin, Ed.\hskip 1em plus 0.5em minus 0.4em\relax
  Bristol: Institute of Physics Publishing, 2001.

\bibitem{buta2011}
------, ``Galaxy morphology,'' in \emph{Planets, Stars, and Stellar Systems},
  T.~D. Oswalt and W.~C. Keel, Eds., 2011, vol.~6, pp. 1--89.

\bibitem{vaucouleurs1959}
G.~De~Vaucouleurs, \emph{Classification and Morphology of External
  Galaxies}.\hskip 1em plus 0.5em minus 0.4em\relax Berlin, Heidelberg:
  Springer Berlin Heidelberg, 1959, pp. 275--310.

\bibitem{petrosian1976}
V.~Petrosian, ``Surface brightness and evolution of galaxies,''
  \emph{Astrophysical Journal}, vol. 209, pp. L1--L5, Oct. 1976.

\bibitem{sersic1963}
J.~L. S\'ersic, ``Influence of the atmospheric and instrumental dispersion on
  the brightness distribution in a galaxy,'' \emph{Boletin de la Asociacion
  Argentina de Astronomia}, vol.~6, p.~41, 1963.

\bibitem{jaffe1983}
W.~Jaffe, ``A simple model for the distribution of light in spherical
  galaxies,'' \emph{Monthly Notices of the Royal Astronomical Society}, vol.
  202, pp. 995--999, 1983.

\bibitem{vanderWel2008}
A.~van~der Wel, ``The morphology-density relation: a constant of nature,'' vol.
  245, 2008, pp. 59--62.

\bibitem{naim1995}
A.~Naim, O.~Lahav, L.~Sodr\'e~Jr., and M.~Storrie-Lombardi, ``Automated
  morphological classification of {APM} galaxies by supervised artificial
  neural networks,'' \emph{Monthly Notices of the Royal Astronomical Society},
  vol. 275, no. 567, 1995.

\bibitem{owens1996}
E.~A. Owens, R.~E. Griffiths, and K.~Ratnatunga, ``Using oblique decision trees
  for the morphological classification of galaxies,'' \emph{Monthly Notices of
  the Royal Astronomical Society}, vol. 281, no. 153, 1996.

\bibitem{bazell2001}
D.~Bazell and D.~W. Aha, ``Ensembles of classifiers for morphological galaxy
  classification,'' \emph{The Astrophysical Journal}, vol. 548, pp. 219--233,
  2001.

\bibitem{delacalleja2004}
J.~De~la Calleja and O.~Fuentes, ``Machine learning and image analysis for
  morphological galaxy classification,'' \emph{Monthly Notices of the Royal
  Astronomical Society}, vol. 349, pp. 87--93, 2004.

\bibitem{goderya2002}
S.~N. Goderya and S.~M. Lolling, ``Morphological classification of galaxies
  using computer vision and {ANN}s,'' \emph{Astrophysics and Space Science},
  vol. 279, pp. 377--387, 2002.

\bibitem{shamir2011}
L.~Shamir, ``Ganalyzer: A tool for automatic galaxy image analysis,'' \emph{The
  Astrophysical Journal}, vol. 736, no.~2, p. 141, 2011.

\bibitem{shamir2012}
------, ``Automatic detection of peculiar galaxies in large datasets of galaxy
  images,'' \emph{Journal of Computational Science}, vol.~3, no.~3, pp.
  181--189, 2012.

\bibitem{dieleman2015}
S.~Dieleman, K.~W. Willett, and J.~Dambre, ``Rotation-invariant convolutional
  neural networks for galaxy morphology prediction,'' \emph{Monthly Notices of
  the Royal Astronomical Society}, vol. 450, pp. 1441--1459, 2015.

\bibitem{goshtasby1985}
A.~Goshtasby, ``Template matching in rotated images,'' \emph{IEEE Trans.
  Pattern Anal. Mach. Intell.}, vol.~7, no.~3, pp. 338--344, May 1985.

\bibitem{hu1962}
M.~K. Hu, ``Visual pattern recognition by moment invariants,'' \emph{IRE Trans.
  Info. Theory}, vol.~8, pp. 179--187, 1962.

\bibitem{flusser2000}
J.~Flusser, ``On the independence of rotation moment invariants,''
  \emph{Pattern Recognition}, vol.~33, pp. 1405--1410, 2000.

\bibitem{zernike1934}
F.~Zernike, ``Beugungstheorie des schneidenver-fahrens und seiner verbesserten
  form, der phasenkontrastmethode,'' \emph{Physica}, vol.~1, pp. 689--704,
  1934.

\bibitem{khotanzad1990}
A.~Khotanzad and Y.~H. Hong, ``Invariant image recognition by {Z}ernike
  moments,'' \emph{IEEE Trans. Pattern Analysis and Machine Intelligence},
  vol.~12, no.~5, pp. 489--497, May 1990.

\bibitem{tang1991}
Y.~Y. Tang, C.~H. D., and C.~Y. Suen, ``Transformation ring projection {(TRP)}
  algorithm and its {VLSI} implementation,'' \emph{International Journal of
  Pattern Recognition and Artificial Intelligence}, vol.~5, no. 1-2, pp.
  25--56, 1991.

\bibitem{grace1991}
A.~E. Grace and M.~Spann, ``A comparison between {Fourier Mellin} descriptors
  and moment based features for invariant object recognition using neural
  networks,'' \emph{Pattern Recog. Lett.}, vol.~12, pp. 635--643, 1991.

\bibitem{sheng1991}
Y.~Sheng and C.~Lejeune, ``Invariant pattern recognition using {Fourier Mellin}
  transforms and neural networks,'' \emph{J. Optics}, vol.~22, no.~5, pp.
  223--228, 1991.

\bibitem{adam2000}
S.~Adam, J.-M. Ogier, C.~Cariou, R.~Mullot, J.~Gardes, and Y.~Lecourtier,
  ``Fourier-mellin based invariants for the recognition of multi-oriented and
  multi-scaled shapes - application to engineering drawings analysis,''
  \emph{Invariants for Pattern Recognition and Classification}, pp. 125--146,
  2000.

\bibitem{adam2000a}
S.~Adam, J.~M. Ogier, C.~Cariou, R.~Mullot, J.~Labiche, and J.~Gardes, ``Symbol
  and character recognition: application to engineering drawings,''
  \emph{International Journal on Document Analysis and Recognition (IJDAR)},
  vol.~3, pp. 89--101, Dec. 2000.

\bibitem{casasent1976}
D.~Casasent and D.~Psaltis, ``Position, rotation, and scale invariant optical
  correlation,'' \emph{Applied Optics}, vol.~15, pp. 1795--1799, 1976.

\bibitem{goh1985}
S.~Goh, ``The {M}ellin transformation: Theory and digital filter
  implementation,'' Ph.D. dissertation, Purdue University, West Lafayette,
  I.N., 1985.

\bibitem{ghorbel1994}
F.~Ghorbel, ``A complete invariant description for gray-level images by the
  harmonic analysis approach,'' \emph{Pattern Recog. Lett.}, vol.~15, pp.
  1043--1051, 1994.

\bibitem{derrode2001}
S.~Derrode and F.~Ghorbel, ``Robust and efficient {Fourier Mellin} transform
  approximations for gray-level image reconstruction and complete invariant
  description,'' \emph{Computer Vision and Image Understanding}, vol.~83, pp.
  57--78, 2001.

\bibitem{otsu1979}
N.~Otsu, ``A threshold selection method from gray-level histograms,''
  \emph{IEEE Trans. Sys. Man. Cyber.}, vol.~9, no.~1, pp. 62--66, 1979.

\bibitem{burt1983}
P.~J. Burt and E.~Adelson, ``The laplacian pyramid as a compact image code,''
  \emph{IEEE Trans. on Communications}, vol.~31, no.~4, Apr. 1983.

\bibitem{fawcett2006}
T.~Fawcett, ``An introduction to {ROC} analysis,'' \emph{Pattern Recog. Lett.},
  vol.~27, pp. 861--874, 2006.

\bibitem{vapnik1998}
V.~Vapnik, \emph{Statistical Learning Theory}.\hskip 1em plus 0.5em minus
  0.4em\relax Wiley, New York, NY, 1998.

\bibitem{kay1992}
D.~J.~C. MacKay, ``Bayesian interpolation,'' \emph{Neural Comput.}, vol.~4,
  no.~3, pp. 415--447, 1992.

\bibitem{draper1998}
N.~R. Draper and H.~Smith, \emph{Applied Regression Analysis}.\hskip 1em plus
  0.5em minus 0.4em\relax Hoboken, NJ: Wiley-Interscience, 1998.

\bibitem{huang2012}
G.~B. Huang, H.~Zhou, X.~Ding, and R.~Zhang, ``Extreme learning machine for
  regression and multiclass classification,'' \emph{IEEE Trans. on Systems,
  Man, and Cybernetics, Part B: Cybernetics}, vol.~42, no.~2, pp. 513--529,
  2012.

\bibitem{bishop2006}
C.~M. Bishop, \emph{Pattern Recognition and Machine Learning}.\hskip 1em plus
  0.5em minus 0.4em\relax Springer Press, 2006.

\bibitem{huang2014}
G.~B. Huang, ``An insight into extreme learning machines: Random neurons,
  random features and kernels,'' \emph{Cognitive Computation}, vol.~6, pp.
  376--390, 2014.

\bibitem{lecun2015}
Y.~LeCun, Y.~Bengio, and G.~Hinton, ``Deep learning,'' \emph{Nature}, vol. 521,
  no. 7553, pp. 436--444, 2015.

\bibitem{kingma2015}
D.~P. Kingma, , and J.~L. Ba, ``Adam : A method for stochastic optimization,''
  in \emph{Proc. of Int. Conf. on Learning Representations (ICLR)}, 2015, pp.
  1--13.

\bibitem{glorot2011}
X.~Glorot, A.~Bordes, and Y.~Bengio, ``Deep sparse rectifier neural networks,''
  in \emph{Proc. of the 14th Int. Conf. on Artificial Intelligence and
  Statistics (AISTATS)}, 2011, pp. 315--323.

\bibitem{aniyan2017}
A.~K. Aniyan and K.~Thorat, ``Classifying radio galaxies with the convolutional
  neural network,'' \emph{The Astrophysical Journal Supplement Series}, vol.
  230, no.~2, 2017.

\bibitem{katebi2019}
R.~Katebi, Y.~Zhou, R.~Chornock, and R.~Bunescu, ``Galaxy morphology prediction
  using capsule networks,'' \emph{Monthly Notices of the Royal Astronomical
  Society}, vol. 486, pp. 1539--1547, 2019.

\bibitem{kuminski2014}
E.~Kuminski, J.~George, J.~Wallin, and L.~Shamir, ``Combining human and machine
  learning for morphological analysis of galaxy images,'' \emph{Publications of
  the Astronomical Society of the Pacific}, vol. 126, pp. 959--967, 2014.

\bibitem{cecotti2016}
H.~Cecotti, ``Active graph based semi-supervised learning using image matching:
  application to handwritten digit recognition,'' \emph{Pattern Recog. Lett.},
  vol.~73, pp. 76--82, 2016.

\end{thebibliography}

\end{document}